\documentclass[]{fairmeta}

\usepackage{amsmath,amssymb}
\usepackage{algorithm}
\usepackage{algorithmic}
\usepackage{colortbl}
\usepackage{listings}
\usepackage{docmute}
\usepackage{xparse}

\definecolor{PromptFrame}{RGB}{0,94,184}
\definecolor{PromptTitle}{RGB}{255,255,255}
\definecolor{SummaryFrame}{RGB}{184,192,204}
\definecolor{SummaryBack}{RGB}{246,248,251}
\definecolor{TableStripe}{gray}{0.96}
\lstdefinelanguage{diff}{
  morecomment=[f][\color{gray}]{@@},
  morecomment=[f][\color{red!70!black}]{-},
  morecomment=[f][\color{green!50!black}]{+}
}

\graphicspath{{Figures/}{analysis/}}
\newcommand{\PaperShortTitle}{\texttt{InferenceEvolve} for Automated Causal Estimators}
  \newcommand{\PaperTitle}{\texttt{InferenceEvolve}: Towards Automated Causal Effect Estimators through Self-Evolving AI}
  \newcommand{\PaperAbstractText}{
Causal inference is central to scientific discovery, yet choosing appropriate methods remains challenging because of the complexity of both statistical methodology and real-world data. Inspired by the success of artificial intelligence in accelerating scientific discovery, we introduce \MethodName{}, an evolutionary framework that uses large language models to discover and iteratively refine causal methods. Across widely used benchmarks, \MethodName{} yields estimators that consistently outperform established baselines: against 58 human submissions in a recent community competition, our best evolved estimator lay on the Pareto frontier across two evaluation metrics. We also developed robust proxy objectives for settings without semi-synthetic outcomes, with competitive results. Analysis of the evolutionary trajectories shows that agents progressively discover sophisticated strategies tailored to unrevealed data-generating mechanisms. These findings suggest that language-model-guided evolution can optimize structured scientific programs such as causal inference, even when outcomes are only partially observed.}

\tcbuselibrary{skins,breakable}

\newcommand{\ArxivMethodsFirst}{}
\newcommand{\ArxivOverviewFirst}{}
\renewcommand{\PaperTitle}{\texttt{InferenceEvolve}: Automated Causal Effect Estimators through Self-Evolving AI}

\newcommand{\MethodName}{{\normalfont\textsc{InferenceEvolve}}}
\makeatletter
\providecommand{\manualsectionlabel}[2]{%
  \phantomsection
  \def\@currentlabel{#2}%
  \def\@currentlabelname{#2}%
  \label{#1}}
\makeatother

\newtcolorbox{promptbox}[2][]{
  breakable,
  enhanced,
  colback=white,
  colframe=PromptFrame,
  colbacktitle=PromptFrame,
  coltitle=PromptTitle,
  fonttitle=\bfseries,
  title={#2},
  boxrule=0.6pt,
  left=6pt,
  right=6pt,
  top=6pt,
  bottom=6pt,
  before skip=8pt,
  after skip=8pt,
  #1
}

\newtcolorbox{summarybox}[2][]{
  breakable,
  enhanced,
  colback=SummaryBack,
  colframe=SummaryFrame,
  fonttitle=\bfseries,
  title={#2},
  boxrule=0.5pt,
  left=6pt,
  right=6pt,
  top=5pt,
  bottom=5pt,
  before skip=8pt,
  after skip=8pt,
  #1
}

\newcommand{\fnm}[1]{#1}
\newcommand{\sur}[1]{ #1}
\newcommand{\orgdiv}[1]{#1}
\newcommand{\orgname}[1]{#1}
\newcommand{\orgaddress}[1]{#1}
\newcommand{\city}[1]{#1}
\newcommand{\postcode}[1]{#1}
\newcommand{\country}[1]{#1}

\let\fairmetatitle\title
\RenewDocumentCommand{\title}{o m}{\fairmetatitle{#2}}

\let\fairmetaauthor\author
\RenewDocumentCommand{\author}{s o m}{%
  \IfValueTF{#2}{\fairmetaauthor[#2]{#3}}{\fairmetaauthor{#3}}%
}

\let\fairmetaaffiliation\affiliation
\NewDocumentCommand{\affil}{s o m}{%
  \IfValueTF{#2}{\fairmetaaffiliation[#2]{#3}}{\fairmetaaffiliation{#3}}%
}

\RenewDocumentCommand{\email}{m}{}

\let\fairmetabibliography\bibliography
\RenewDocumentCommand{\bibliography}{m}{\fairmetabibliography{#1}}

\begin{document}

\bibliographystyle{plainnat}
\setcitestyle{authoryear,round,aysep={,},yysep={;},notesep={; }}

\title[\PaperShortTitle]{\PaperTitle}

\author[1]{\fnm{Can} \sur{Wang}}
\email{cwang271@jh.edu}

\author[1]{\fnm{Hongyu} \sur{Zhao}}
\email{hzhao59@jhu.edu}

\author*[1,2]{\fnm{Yiqun} \sur{Chen}}
\email{yiqunc@jhu.edu}

\affil[1]{\orgdiv{Department of Biostatistics}, \orgname{Johns Hopkins Bloomberg School of Public Health}, \orgaddress{\city{Baltimore}, \postcode{MD 21205}, \country{USA}}}

\affil[2]{\orgdiv{Department of Computer Science}, \orgname{Johns Hopkins University}, \orgaddress{\city{Baltimore}, \postcode{MD 21218}, \country{USA}}}

\abstract{\PaperAbstractText}

\maketitle

\newcommand{\OverviewFigureBlock}{
\begin{figure}[htbp!]
\centering
\includegraphics[width=\textwidth]{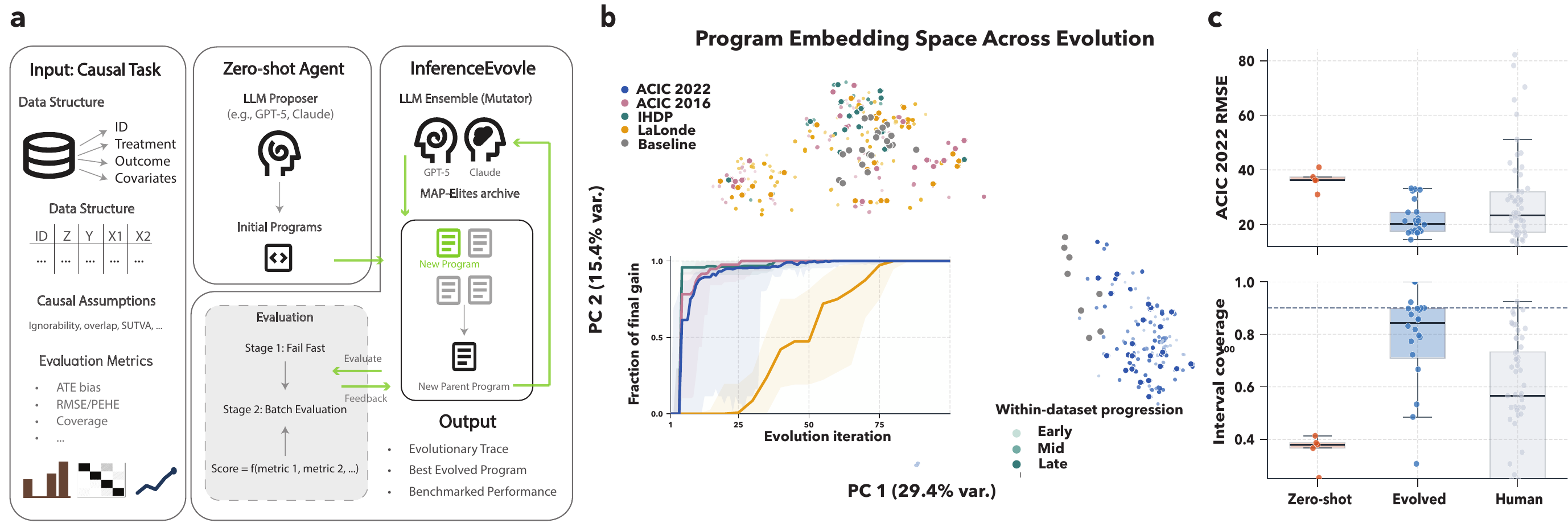}
\caption{\textbf{Overview of \MethodName{} for causal estimator discovery.} \textbf{a}, Schematic of the workflow: a causal task specification seeds a zero-shot program, which is then iteratively improved by an LLM ensemble under benchmark-based feedback, yielding an evolutionary trace and a final estimator. \textbf{b}, Principal-component view of the programs in the OpenAI's \texttt{text-embedding-3-large} embedding space across evolutions. Colors denote benchmarks (gray points denote zero-shot baselines), and darker points indicate later checkpoints within each benchmark. The inset line graph (no points obscured) summarizes normalized best-so-far progress over evolution iterations. \textbf{c}, ACIC~2022~\citep{thal_causal_2023} comparison against zero-shot programs and 58 human competition submissions. Top, RMSE distributions. Bottom, empirical 90\% interval coverage distributions. Across the full distribution, evolved programs improve over zero-shot generation and compare favorably with human submissions.}
\label{fig:overview}
\end{figure}
}

\ifdefined\ArxivOverviewFirst
\OverviewFigureBlock
\fi



\section*{Introduction}

Causal inference answers whether and how an intervention affects an outcome and has been central to scientific discovery and evidence-based decision-making~\citep{hernan2010causal,pearl2009causal,bailey2024causal}. One of the fundamental challenges in causal inference is the \emph{counterfactual problem}: for any unit receiving treatment, we cannot observe what would have happened under control, and vice versa~\citep{imbens2015causal}. Practitioners must therefore rely on assumptions and statistical methods to estimate treatment effects from observational data, where treatment assignment may depend on observed and unobserved confounders~\citep{rosenbaum1983central}. Despite decades of methodological progress---from propensity score methods~\citep{stuart2010matching,austin2011introduction} to doubly robust estimators~\citep{chernozhukov2018double,kennedy2024semiparametric,bang2005doubly} and causal machine learning~\citep{feuerriegel2024causal}---two persistent challenges remain. First, \emph{which estimator to use}: the landscape of available methods is vast and growing, and selecting the right approach requires substantial expertise about both the statistical properties of estimators and the substantive features of the data at hand. Second, \emph{how to evaluate estimators}. Since ground-truth treatment effects are unobservable in real data, researchers routinely rely on semi-synthetic benchmarks with known data-generating processes to assess and compare methods~\citep{hill_bayesian_2011,dorie2019automated,thal_causal_2023}.

To illustrate with a concrete example, consider estimating the effect of a job training program on subsequent earnings. Each individual $i$ has potential outcomes $Y_i(1)$ and $Y_i(0)$ under treatment and control, but only one is observed: $Y_i = Z_i Y_i(1) + (1-Z_i)Y_i(0)$, where $Z_i$ is the treatment indicator. Stakeholders are often interested in the average treatment effect $\text{ATE} = \mathbb{E}[Y(1) - Y(0)]$, which requires estimating the missing potential outcomes. A human expert would typically proceed by: identifying the causal estimand under specific assumptions; choosing an estimation strategy (e.g., inverse propensity weighting, outcome regression, or doubly robust methods); selecting the models used to estimate nuisance parameters; and constructing confidence intervals. Each step involves modeling choices that can substantially affect the final estimate. To evaluate the quality of this pipeline, researchers often use semi-synthetic datasets that mimic the real-world settings they care about, and compute quantitative metrics such as bias, efficiency, and calibration of uncertainty~\citep{imbens2015causal}.

The causal inference workflow is well-suited to automation for two reasons. First, the space of causal estimation strategies is combinatorially large but structured---different choices of identification strategy, nuisance models, and aggregation schemes can be composed programmatically. Second, the standard evaluation paradigm in causal inference already relies on simulated datasets with known ground truth, providing a natural fitness signal for optimization. Recent work has demonstrated the potential of large language model (LLM)-based agents for scientific tasks~\citep{gao2024empowering,gottweis2025towards,yamada2025ai,swanson2025virtual}. In parallel, evolutionary code generation frameworks such as Autoresearch~\citep{karpathy_autoresearch_2026}, AlphaEvolve~\citep{novikov2025alphaevolve} and OpenEvolve~\citep{openevolve} have shown that LLMs can discover state-of-the-art algorithms through iterative program improvement guided by automated evaluation. By contrast, existing LLM work in causality has largely focused on \emph{individual components}: whether LLMs can infer directional causality from text, select among pre-specified causal workflows~\citep{acharya_causcibench_2025,bazgir_causal_2025}, or augment existing causal analyses. To our knowledge, open-ended estimator discovery---where the agent must search jointly over identification strategy, nuisance models, weighting or meta-learning structure, and interval construction---has not previously been studied in causal inference.

Motivated by this gap, we introduce \MethodName{}, an evolutionary framework that treats causal inference estimators as code programs, evaluates them on benchmark datasets, and iteratively refines them through LLM-guided mutations (Figure~\ref{fig:overview}). Our work makes the following contributions. First, we introduce an open-ended LLM-agent framework for causal estimator discovery across four diverse benchmarks: ACIC~2022~\citep{thal_causal_2023}, ACIC~2016~\citep{dorie2019automated}, IHDP~\citep{hill_bayesian_2011}, and LaLonde~\citep{lalonde1986evaluating,dehejia1999causal}, which span panel data, cross-sectional, semi-synthetic, and observational study designs. Because these benchmarks expose many semi-synthetic or competition replicates rather than a single fixed table, each run can optimize on one subset of replicates and report results only on disjoint held-out replicates. This does not eliminate possible benchmark memorization by frontier models, but it does force gains beyond the zero-shot seed to generalize across unseen replicate draws within the benchmark. Second, we develop proxy evaluation metrics based on doubly robust estimators of the treatment effect, which enable iterative refinement without access to ground-truth counterfactuals. Our proof of concept provides a path toward causal estimator development when true counterfactuals are unavailable. Third, we empirically examine the distribution of evolved estimator performance across different experimental conditions and provide practical guidance for researchers designing system parameters for this framework. More broadly, we use these benchmarks as a controlled testbed for automated method discovery: the search objective, held-out evaluation, and post hoc program analysis are all explicit and auditable. We also characterize the qualitative refinement steps, method compositions, and code-similarity structure of these programs beyond the final quantitative scores, offering domain insight that goes beyond the final metric alone.

\ifdefined\ArxivOverviewFirst\else
\OverviewFigureBlock
\fi


\long\def\ResultsSectionContent{
\section*{Results}
We first compared the evolved estimators against quantitative baselines, including the zero-shot LLM baseline, off-the-shelf causal inference methods across all four benchmarks, and, for ACIC~2022, the distribution of 58 human competition submissions. We then qualitatively examined the learned programs, focusing on the estimator families that emerged at convergence, their overlap with existing work, and their coding and conceptual complexity.

\subsection*{\MethodName{} discovers strong causal estimators}
For the full true-score sweep, each dataset has 24 evolved final programs from a $2 \times 3 \times 2 \times 2$ factorial design over two primary LLMs from the GPT and Gemini families (GPT-5 and Gemini~3~Flash~Preview; \citealp{achiam2023gpt,team2024gemini}), regularization weight $\lambda$ (larger $\lambda$ puts more weight on ATE estimation error and coverage error), held-out split (20\% versus 50\% training), and replicate seed. All main-text figures and tables focus on the prespecified 20:80 split; the matched 50:50 reruns are used only as sensitivity analyses in the Supplementary Information. Search received feedback only from the training replicates, whereas every quantitative claim in the main text is evaluated on the untouched held-out replicates. Proxy-guided runs were analyzed separately under the same benchmarks with search-time access only to the proxy metrics as opposed to the ground truth counterfactuals.

On the ACIC~2022 data with rich human competition results ($n=58$), the 12 final estimators from different configurations of \MethodName{} obtained a median root mean squared error (RMSE) 18.8 and median empirical coverage 0.84 (target level, 0.9), compared with 23.2 and 0.57, respectively, across the 58 human submissions (Fig.~\ref{fig:overview}). By RMSE, 8 of the 12 evolved final estimators outperformed the median human submission and the best evolved run (RMSE 14.4) outperformed 51 of 58 submissions. The top two evolved programs also lie on the \emph{Pareto frontier} with respect to RMSE and absolute deviation from nominal coverage~\citep{miettinen1999nonlinear}; that is, \emph{no human submission achieved both lower RMSE and coverage closer to the target 90\% level than either of these two programs}. Without the self-evolving improvement, LLM performance was clearly weaker: for example, GPT-5 zero-shot programs produced a difference-in-differences estimator with RMSE 33.1 and coverage 0.40 (Supplementary Tables~\ref{tab:ext-acic-zeroshot} and \ref{tab:ext-ihdp-zeroshot}).

For the other semi-synthetic datasets without direct human benchmarks, we compared final programs against commonly used reference estimators spanning regression adjustment, propensity weighting, causal forests, tree-based meta-learners, and benchmark-specific panel baselines (see Methods). We display the held-out test-set metrics across all configurations and runs in Fig.~\ref{fig:baseline}, with detailed metrics of the best runs reported in Table~\ref{tab:baseline}. Across all benchmarks, \MethodName{} improved held-out performance relative to the strongest fixed package baselines. On IHDP, the strongest baseline (BART) achieved $\sqrt{\mathrm{PEHE}}$ 2.41 and ATE error 0.11, compared with 1.22 and 0.06 for the best true-score run; the best proxy-guided run reached a lower $\sqrt{\mathrm{PEHE}}$ of 1.034 while yielding ATE error 0.074. On ACIC~2016, the strongest baseline (causal forest) achieved 1.28 and 0.15, compared with 0.86 and 0.09 for the best \MethodName{} run. On LaLonde, the strongest package baseline (BART) achieved 0.77 for $\sqrt{\mathrm{PEHE}}$ and 0.07 for ATE, while the design-based difference-in-means reference achieved normalized ATE error 0.05; the best true-score run reached 0.598 and 0.033, whereas the best proxy-guided run reached 0.693 and 0.054, improving over the package $\sqrt{\mathrm{PEHE}}$ baseline but not over the stronger design-based ATE comparator. These paired $\sqrt{\mathrm{PEHE}}$/ATE comparisons show that the gains were not limited to average calibration: under true-score guidance the evolved estimators also improved unit-level heterogeneity recovery on all three individual treatment effect benchmarks.

The final programs produced by \MethodName{} tended to be longer in code length and more complex in terms of methodology than the off-the-shelf baselines. This increase in complexity should be interpreted together with the evaluation protocol: the search never saw the held-out metrics reported in Fig.~\ref{fig:baseline} and Table~\ref{tab:baseline}; those numbers were computed only after the final program was frozen. When GPT-5 was used to transcribe the final code into manuscript-style methods paragraphs, the resulting descriptions were consistently shortest for the baseline programs and longer for the true-evolved and proxy-evolved programs across all four datasets; the same ordering was also evident in code line counts (Fig.~\ref{fig:baseline}). This pattern was robust to the choice of transcription LLM, including Gemini~2.5~Pro and Claude~Sonnet~4.5. Full distributions of code length, measured in characters and lines, are shown in Supplementary Figs.~\ref{fig:ext-method-length} and \ref{fig:ext-code-length}.

\begin{figure}[htbp!]
\centering
\includegraphics[width=\textwidth]{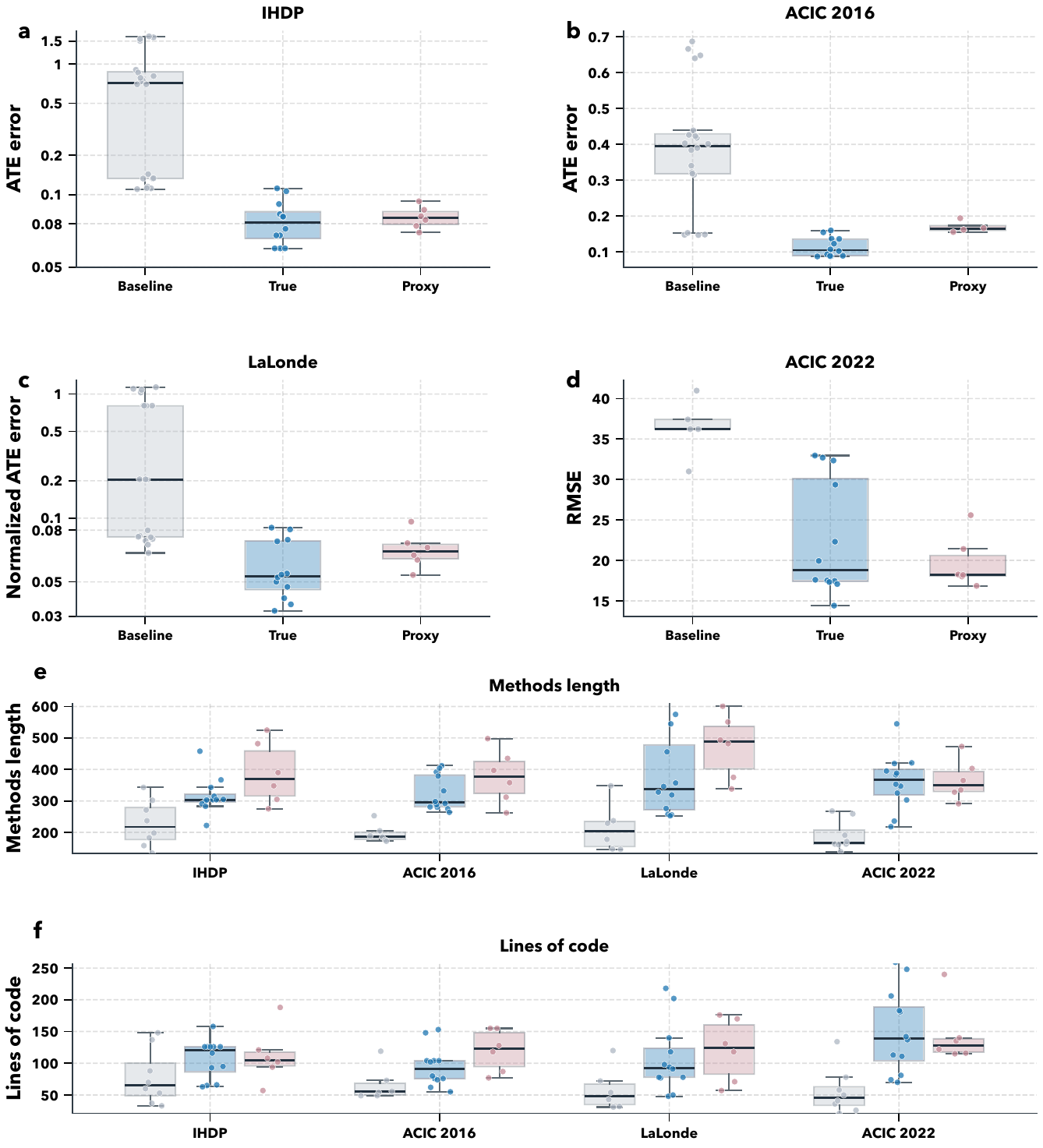}
\caption{\textbf{Held-out performance distributions for baselines, true-evolved programs, and proxy-evolved programs, with text- and code-complexity summaries.} Panels a--d show the metric distributions on the held-out set for all baseline methods and the final true-evolved and proxy-evolved runs. Panels e and f summarize GPT-5 methods length and lines of code for different programs across datasets.}
\label{fig:baseline}
\end{figure}

\begin{table}[t]
\centering
\caption{\textbf{Best evolved estimators versus baseline methods.} The baseline column gives the candidate used for each row: the strongest package baseline in Fig.~\ref{fig:baseline}. \emph{Best evolved} and \emph{best $\Delta$ (\%)} refer to the single best final program under each guidance mode, where positive \emph{best $\Delta$} indicates percent error reduction relative to the row-specific baseline comparator. \emph{Mean $\pm$ SE} and \emph{\% beat} summarize the performance across all parameter settings for \MethodName{}.}
\label{tab:baseline}
\scriptsize
\setlength{\tabcolsep}{2pt}
\begin{tabular}{llccccc}
\toprule
Dataset & Baseline & Mode & Best & Mean $\pm$ SE & \% beat & Best $\Delta$ (\%) \\
\midrule
\shortstack[l]{IHDP\\$\sqrt{\mathrm{PEHE}}$} & 2.411 (BART) & True & 1.222 & $1.392 \pm 0.021$ & 100\% & 49.3\% \\
 &  & Proxy & \textbf{1.034} & $1.632 \pm 0.311$ & 83\% & 57.1\% \\
\shortstack[l]{\\ATE error} & 0.110 (BART) & True & \textbf{0.063} & $0.081 \pm 0.005$ & 100\% & 42.9\% \\
 &  & Proxy & 0.074 & $0.084 \pm 0.003$ & 100\% & 33.0\% \\
\midrule
\shortstack[l]{ACIC~2016\\$\sqrt{\mathrm{PEHE}}$} & 1.283 (Causal forest) & True & \textbf{0.858} & $1.011 \pm 0.042$ & 100\% & 33.1\% \\
 &  & Proxy & 1.504 & $2.037 \pm 0.196$ & 0\% & $-$17.2\% \\
\shortstack[l]{\\ATE error} & 0.147 (Causal forest) & True & \textbf{0.087} & $0.114 \pm 0.011$ & 100\% & 40.6\% \\
 &  & Proxy & 0.156 & $0.170 \pm 0.008$ & 0\% & $-$5.7\% \\
\midrule
\shortstack[l]{LaLonde\\$\sqrt{\mathrm{PEHE}}$} & 0.765 (BART) & True & \textbf{0.598} & $0.704 \pm 0.016$ & 83\% & 21.8\% \\
 &  & Proxy & 0.693 & $0.754 \pm 0.017$ & 67\% & 9.4\% \\
\shortstack[l]{ATE} & 0.048 (Diff-in-means) & True & \textbf{0.033} & $0.057 \pm 0.004$ & 17\% & 30.8\% \\
 &  & Proxy & 0.054 & $0.070 \pm 0.005$ & 0\% & $-$12.6\% \\
\midrule
\shortstack[l]{ACIC~2022\\RMSE} & 23.40 (Backdoor adj.) & True & \textbf{14.41} & $22.58 \pm 2.39$ & 50\% & 38.4\% \\
 &  & Proxy & 16.85 & $19.72 \pm 1.33$ & 83\% & 28.0\% \\
\bottomrule
\end{tabular}
\end{table}

\begin{figure}[htbp!]
\centering
\includegraphics[width=\textwidth]{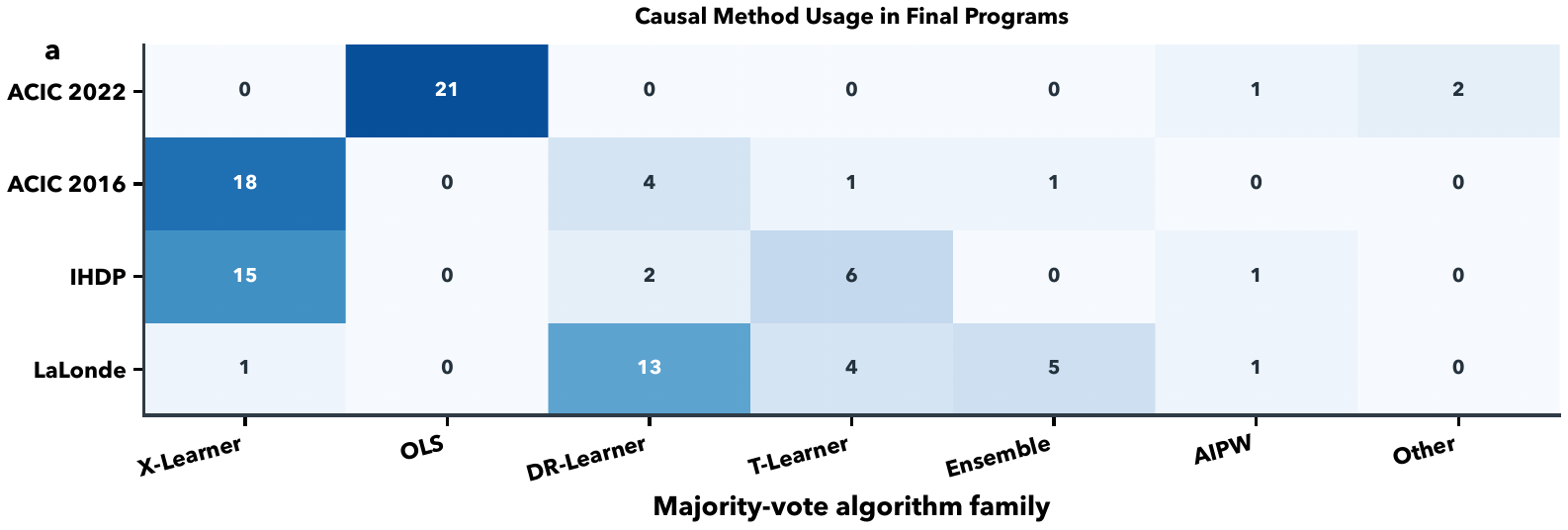}

\vspace{0.75em}
\includegraphics[width=\textwidth]{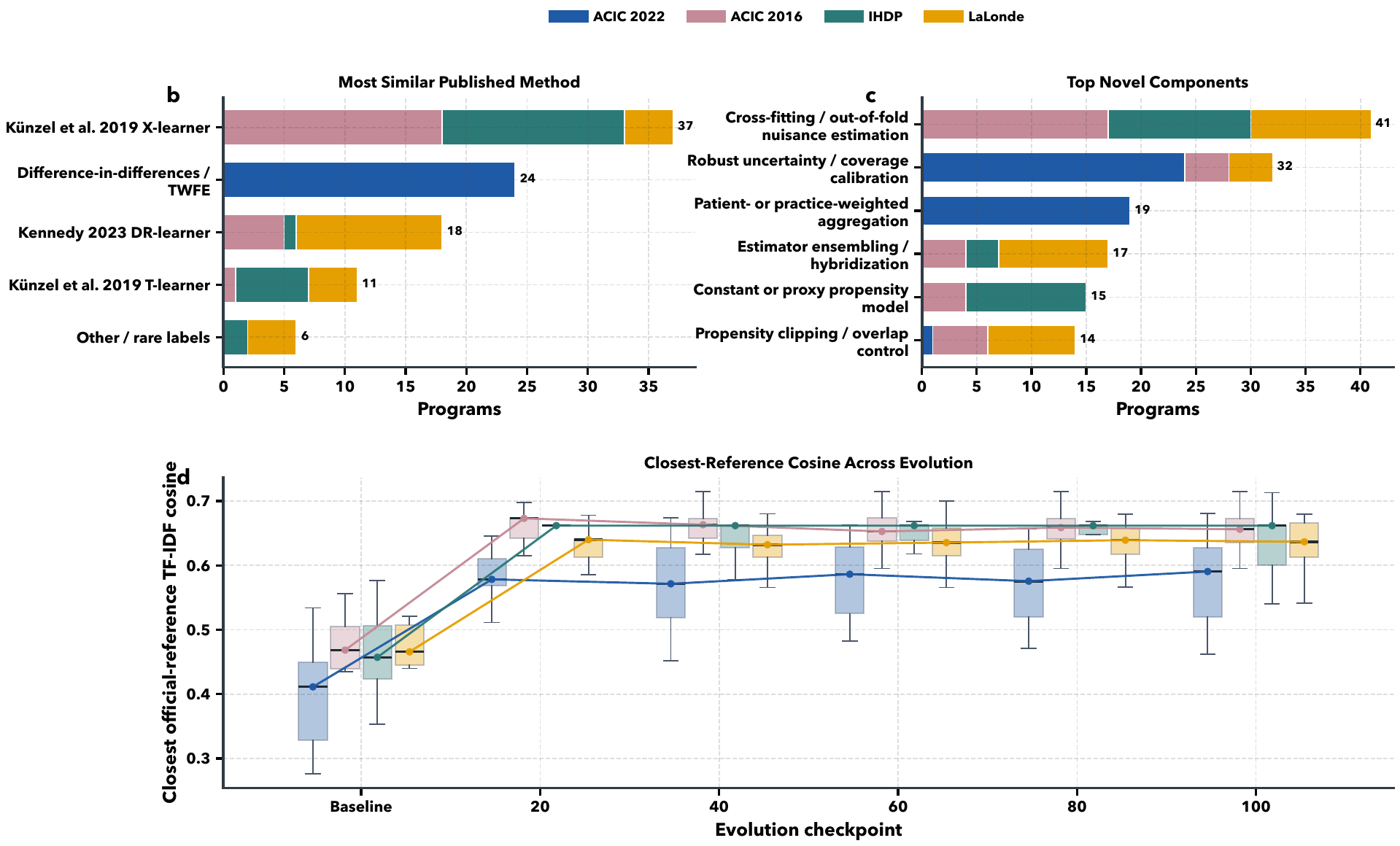}
\caption{\textbf{\MethodName{} converges to dataset-specific estimator families without collapsing onto a single public template.} \textbf{a}, Majority-vote-based algorithm-family assignments for the 96 final evolved programs; each cell reports the number of final programs from that dataset assigned to that family. \textbf{b}, Majority-vote-based most similar published method. \textbf{c}, Novel algorithmic components extracted from the union of LLM-as-a-judge reports for each program. \textbf{d}, Tokenized TF-IDF cosine similarity between each program and its closest official reference wrapper across zero-shot baselines and checkpoints, with boxplots stratified by dataset.}
\label{fig:structure}
\end{figure}

\subsection*{Evolution discovers dataset-specific and novel estimator families}

Beyond held-out performance, it is also important to ask whether evolution reveals a structured search process rather than generic code drift. We summarize our findings in Figure~\ref{fig:structure}: 1. different benchmarks converge to different estimator families; and 2. the evolved programs can be anchored to a set of benchmark-specific published methods and refinements. 

We first note that the converged pool of estimators were strongly dataset-specific, which explains the distinct clusters in PCA plot in Figure~\ref{fig:overview}: since it is a panel data with the same units across timepoints, ACIC~2022 concentrated almost entirely on OLS / panel estimators (21/24). By contrast, ACIC~2016 converged on X-learners~\citep{kunzel2019metalearners} (18/24), IHDP on X- and T-learners~\citep{kunzel2019metalearners} (15/24 and 6/24), and LaLonde on DR-learner~\citep{kennedy2023towards} and ensemble hybrids (13/24 and 5/24). Across all 96 final programs, the majority-vote family counts were X-learner (34), OLS (21), DR-learner (19), and T-learner (11). The three judges in the LLM-as-a-judge pipeline agreed on the coarse family label for 84/96 programs, but agreed on a single nearest named published method for only 42/96, indicating convergence at the level of method families and components rather than exact textbook replication.

Closest official-reference TF-IDF cosine increased from an overall median of 0.446 among the zero-shot baseline programs to 0.640 by checkpoint~20, then remained nearly flat through checkpoint~100 (median 0.638). By checkpoint~100, the dataset-specific medians were 0.591 for ACIC~2022, 0.656 for ACIC~2016, 0.662 for IHDP, and 0.637 for LaLonde. This pattern is consistent with rapid movement toward familiar estimator families followed by benchmark-specific refinement. Relative to the closest official references, the most common novel components were cross-fitting or out-of-fold nuisance estimation (41/96 programs), robust uncertainty or coverage calibration (32/96), patient- or practice-weighted aggregation (19/96), estimator ensembling or hybridization (17/96), constant or proxy propensity models (15/96), and propensity clipping or overlap control (14/96). Pairwise code similarity among evolved programs again showed dataset-specific convergence: overall TF-IDF similarity was higher within datasets than between datasets (0.359 versus 0.228), and the same ordering held for \texttt{text-embedding-3-large} embeddings (0.776 versus 0.665); see Supplementary Section~\ref{sec:supp-program-analysis}.

\begin{summarybox}{Program-analysis highlights}
\textbf{Judge agreement.} The three judges in the LLM-as-a-judge pipeline agreed on the coarse algorithm family for 84 of 96 final programs, but on a single nearest published method for only 42 of 96.\par\smallskip
\textbf{Reference trajectory.} Closest-reference TF-IDF cosine rose from 0.446 in the zero-shot baselines to 0.640 by checkpoint~20, then stayed near 0.64 through checkpoint~100 rather than continuing to collapse toward a single template.\par\smallskip
\textbf{Recurring refinements.} Cross-fitting, uncertainty calibration, weighting, and ensembling were the most repeated component-level changes across the evolved programs.
\end{summarybox}

These structural shifts were accompanied by larger programs and longer scientific write-ups. Across datasets, baseline programs had a median size of 54 lines and 2{,}004 characters, compared with 104 lines and 3{,}922 characters for true-evolved programs and 119.5 lines and 4{,}698.5 characters for proxy-evolved programs (Supplementary Fig.~\ref{fig:ext-code-length}). Translating each program into a manuscript-style methods paragraph with GPT-5, Gemini~2.5~Pro, or Claude~Sonnet~4.5 yielded the same ordering: median description length increased from 172.5 words for baselines to 263.0 for true-evolved programs and 279.5 for proxy-evolved programs (Fig.~\ref{fig:baseline}). Table~\ref{tab:solution_families} summarizes this qualitative picture and situates each dataset-specific family relative to the closest official reference wrapper; the content has been reviewed by two graduate students with causal inference expertise to filter for potential hallucination and bias from the LLM-as-a-judge pipeline~\citep{zheng2023judging}.

\begin{table}[htbp!]
\centering
\caption{\textbf{Representative solution families discovered by evolution.} Final programs cluster into dataset-specific estimator families with recognizable refinement patterns. Similarity values report mean within-dataset TF-IDF cosine similarity among final programs ($\pm$ s.d.) and the median closest official-reference embedding cosine. The ``closest official reference'' column should be interpreted as the nearest single wrapper in embedding space rather than as a second family label: for example, ACIC~2022's weighting-heavy panel estimators sit nearest the IPW wrapper, whereas IHDP's tree- and forest-based nuisance structure pulls that benchmark closest to the causal-forest wrapper.}
\label{tab:solution_families}
\begingroup
\footnotesize
\setlength{\tabcolsep}{3.2pt}
\renewcommand{\arraystretch}{1.24}
\begin{tabular}{
  >{\raggedright\arraybackslash}p{1.3cm}
  >{\raggedright\arraybackslash}p{2.05cm}
  >{\raggedright\arraybackslash}p{3.7cm}
  >{\centering\arraybackslash}p{1.45cm}
  >{\raggedright\arraybackslash}p{1.65cm}
  >{\centering\arraybackslash}p{1.35cm}
}
\toprule
\multicolumn{1}{c}{Dataset} & \multicolumn{1}{c}{\shortstack[c]{Dominant\\family}} & \multicolumn{1}{c}{\shortstack[c]{Typical refinements\\during evolution}} & \multicolumn{1}{c}{\shortstack[c]{Within-dataset\\TF-IDF}} & \multicolumn{1}{c}{\shortstack[c]{Closest official\\reference}} & \multicolumn{1}{c}{\shortstack[c]{Median closest-ref.\\emb.\ cosine}} \\
\midrule
ACIC~2022 & TWFE / DiD / panel fixed-effects estimators & Pre-period anchoring, detrending, patient weighting, clustered or small-sample inference, and coverage-oriented calibration & $0.336 \pm 0.110$ & IPW & 0.585 \\
\rowcolor{TableStripe}
ACIC~2016 & Meta-learners with boosting and DR correction & Cross-fitting, propensity clipping, one-hot encoding of mixed covariates, X-learner or DR-learner structures, and AIPW-style aggregation & $0.353 \pm 0.173$ & DR-learner & 0.678 \\
IHDP & T-/X-learners with boosted or forest nuisance models & Outcome-model tuning, DR pseudo-outcome refinement, ATE calibration, and more stable nuisance estimation & $0.415 \pm 0.244$ & Causal forest & 0.691 \\
\rowcolor{TableStripe}
LaLonde & Weighting and DR hybrids for observational bias correction & Propensity weighting, AIPW/DR correction, adaptive regularization, and ensemble smoothing under small-sample instability & $0.331 \pm 0.116$ & DR-learner & 0.707 \\
\bottomrule
\end{tabular}
\endgroup
\end{table}

\subsection*{Search dynamics differ across models and regularization settings}

\begin{figure}[htbp!]
\centering
\includegraphics[width=\textwidth]{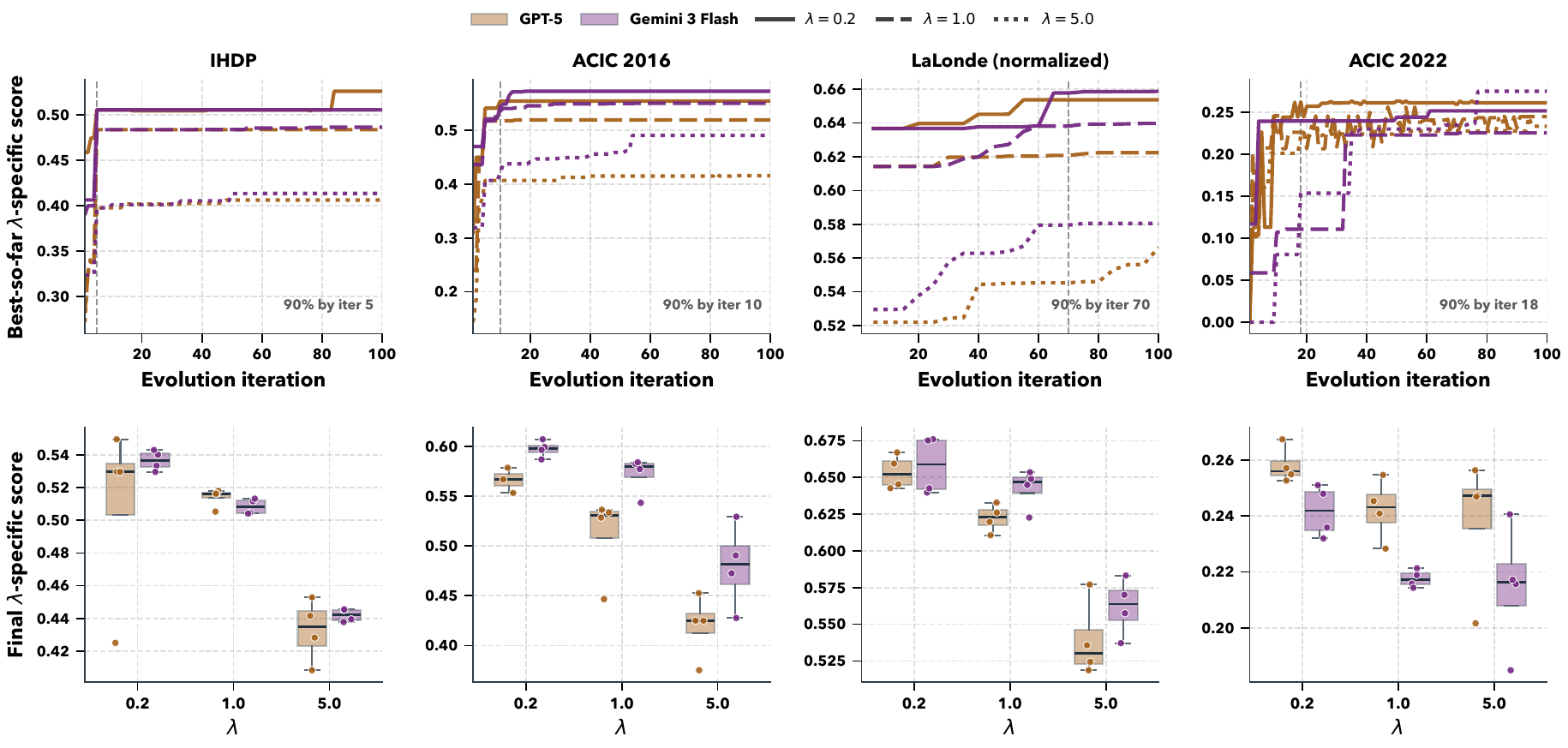}
\caption{\textbf{Search dynamics and sensitivity to the regularization weight $\lambda$.} Top row: median best-so-far search-score trajectories over 100 evolution iterations for the 20:80 main-analysis runs, stratified by primary model and $\lambda$. Color denotes model and line style denotes $\lambda$; dashed lines mark the median iteration at which runs first reached 90\% of their final best-so-far score. Bottom row: final search-score distributions for the same 20:80 settings, pooling over replicate seeds within each $\lambda$--model cell. Because $\lambda$ enters the scalarized score itself, the bottom-row values should be interpreted as within-setting search behavior rather than as directly comparable cross-$\lambda$ measures of estimator quality.}
\label{fig:dynamics-lambda}
\end{figure}

In addition to the endpoints, we look into the evolution dynamics across datasets: the median run reached 90\% of its final best-so-far score by iteration~5 on IHDP, iteration~12 on ACIC~2016, and iteration~16 on ACIC~2022, whereas LaLonde improved more gradually and reached the same threshold around iteration~75. On ACIC~2022, the strongest run moved from an under-calibrated weighted TWFE seed to an augmented DiD estimator by adding pre-period anchoring, post-period differencing, trend adjustment, and cluster-robust inference, reducing RMSE from 33.1 at initialization to 15.2 at its final best checkpoint; the single best-RMSE final run reported in Table~\ref{tab:baseline} reached 14.41. On IHDP and ACIC~2016, the search similarly progressed from simple T- or X-learner structures toward more stable meta-learners with better nuisance estimation and doubly robust aggregation.

Figure~\ref{fig:dynamics-lambda} also separates the two primary models, GPT-5 and Gemini~3~Flash~Preview, and shows that the effect of $\lambda$ is broadly similar across them. Smaller $\lambda$ generally favored lower held-out $\sqrt{\mathrm{PEHE}}$ and/or ATE error on IHDP, ACIC~2016, and normalized LaLonde, whereas ACIC~2022 was comparatively flat over the tested range. Because $\lambda$ changes the scalarized search objective itself, these cross-$\lambda$ conclusions are based on the held-out component metrics rather than on the absolute bottom-row search-score values.

\subsection*{Proxy-guided optimization}

A central difficulty in causal estimator discovery is that the true reward is usually unobserved in real data. We therefore replaced hidden-truth objectives with observed-data proxy rewards built from cross-fitted doubly robust pseudo-outcomes for IHDP, ACIC~2016, and normalized LaLonde, and from a DR difference-in-differences pseudo-target for ACIC~2022 (see Methods). This lets the same evolutionary loop operate even when ground-truth counterfactuals are unavailable.

Proxy fidelity was benchmark-dependent but informative. On matched final runs, proxy-to-true combined score correlation was strongest on IHDP ($r = 0.97$), lower on normalized LaLonde ($r = 0.77$) and ACIC~2016 ($r = 0.50$), and for ACIC~2022 the DR-DID hit-rate component tracked true coverage very closely across checkpoint reevaluations ($r = 0.99$). Best proxy-guided estimators still outperformed the reference baselines on IHDP and ACIC~2022, and on LaLonde they improved over the strongest package $\sqrt{\mathrm{PEHE}}$ baseline even though they no longer beat the stronger design-based ATE comparator under the 20:80 main split (Table~\ref{tab:baseline}). ACIC~2016 remained the clearest failure case, highlighting that optimal proxy design remains an open problem rather than a solved component.
}


\long\def\DiscussionSectionContent{
\section*{Discussion}

With the growing scientific and coding capabilities of language models, their use in causal inference has emerged as a promising direction for automating and improving empirical causal analysis across scientific domains. In this work, we introduced \MethodName{}, a knowledge- and feedback-guided evolutionary framework that steers causal inference methods toward better empirical performance. Across widely used semi-synthetic benchmarks, our evolutionary coding approach proved highly effective: under the prespecified 20:80 main split, it beat the strongest fixed package baseline on every IHDP and ACIC~2016 setting, on most normalized LaLonde settings, and on several ACIC~2022 settings, while remaining competitive even when using a proxy metric to evaluate unobserved effects. Beyond raw performance, our inspection pipeline combines LLM-derived and human-verified insights from evolutionary traces, final-program characterizations, code-similarity analyses, and code-to-manuscript translations, providing scientific interpretability to the search process. Rather than producing only opaque score improvements, \MethodName{} discovers structured, dataset-specific families of solutions. Overall, our results highlight a reusable template for automated methodological search in scientific code spaces, not just a benchmark-specific prompt-engineering exercise.

From a broader AI perspective, we view these results as evidence for a more general capability of LLM-guided evolutionary search: not merely fitting predictors, but rediscovering and adapting scientific procedures. In our setting, the agent is not optimizing a single supervised objective over a fixed hypothesis class. Instead, it must navigate an open-ended design space involving identification assumptions, nuisance estimation, aggregation strategies, and uncertainty quantification, while adapting these choices to different data-generating regimes that are never directly revealed and can only be inferred indirectly through metric feedback. The broader computational contribution is therefore the full experimental workflow: replicated benchmark draws, strict train-versus-held-out separation, auditable executable programs, and post hoc structural analysis of the discovered methods. Our results provide further evidence that LLMs can help explore structured methodological spaces that have traditionally required expert manual iteration. Another contribution of our work is the use of proxy evaluation metrics to enable evolution without access to ground-truth counterfactuals. Proxy-guided estimators outperformed the best baselines on three of the four benchmarks, showing that ground-truth-free estimator search is feasible, though optimal proxy metric design remains an important open problem.

As with all benchmark-driven work in causal inference, our results are subject to a simulation-to-reality gap: strong performance on semi-synthetic data does not guarantee comparable performance in complex real-world observational studies. Importantly, our plug-and-play design makes it easy to evaluate new benchmarks and real datasets as more sophisticated and practically meaningful resources become available~\citep{hill_bayesian_2011,dorie2019automated,thal_causal_2023}. We therefore view \MethodName{} primarily as an automated estimator-search system within established benchmark regimes, rather than as a drop-in replacement for expert causal analysis in unconstrained real-world settings. Likewise, our baseline comparison is against strong representative package baselines from modern causal ML practice, not an exhaustive benchmark-specific AutoML or hyperparameter-tuning sweep; future work should compare directly against tuned stacking pipelines and causal AutoML systems. In addition, much recent work has emphasized causal discovery and the evaluation of plausible causal relationships~\citep{wang2024causalbench,acharya_causcibench_2025,miliani2025explica,ban2025integrating,shen2025exploring}, and we view \MethodName{} as complementary to those developments: it can naturally be combined with advances in causal discovery as an upstream component of a broader end-to-end pipeline.

Finally, while our benchmark suite covers a useful range of commonly used data-generating settings and difficulty levels, there remains little benchmark infrastructure for multimodal causal inference---settings in which inputs may include images, text, or video, and the goal is to estimate the causal effects of these variables as exposures, confounders, or both. This is an important direction for future work, along with studying the robustness of evolved estimators under realistic distribution shift and missingness, both of which are common in scientific and medical applications. More broadly, it would be valuable to explore whether this approach extends to a wider range of data science tasks, especially beyond settings where evolution is driven by a single scalar outcome and where open-ended methodological search has received little attention.
}


\long\def\MethodsSectionContent{
\section*{Methods}
\subsection*{Benchmark datasets}

We evaluated four benchmark families spanning panel-data estimation, semi-synthetic individual treatment effect estimation, and observational bias correction. For an individual $i$, we use $\tau_i = Y_i(1)-Y_i(0)$ to denote the true individual treatment effect and let $\hat{\tau}_i$ denote its estimate. Following standard performance reporting, we use PEHE (Precision in Estimation of Heterogeneous Effect):
\begin{equation}
\sqrt{\mathrm{PEHE}} =
\sqrt{\frac{1}{N}\sum_{i=1}^{N}(\hat{\tau}_i - \tau_i)^2}
= \frac{\lVert \hat{\boldsymbol{\tau}} - \boldsymbol{\tau} \rVert_2}{\sqrt{N}},
\label{eq:pehe-def}
\end{equation}
and the absolute ATE error
\begin{equation}
|\text{ATE error}| =
\left|\frac{1}{N}\sum_{i=1}^{N}\hat{\tau}_i - \frac{1}{N}\sum_{i=1}^{N}\tau_i\right|.
\label{eq:ate-def}
\end{equation}
In words, PEHE emphasizes the more granular unit-level heterogeneity, whereas ATE error compares only the average treatment effect and can be small even when the individual treatment effects are imperfectly recovered.

\textbf{ACIC 2022 (Track~2).} ACIC~2022 is a longitudinal benchmark motivated by health policy evaluation, with treatment assigned at the practice level and outcomes observed in practice--year panels~\citep{thal_causal_2023}. Each dataset contains the outcome $Y$, treatment indicator $Z$, a post-period indicator, patient counts $n_{jt}$, nine practice-level covariates, and seven aggregated patient covariates. The target estimand is the patient-weighted sample average treatment effect on the treated (SATT) over the post-intervention years:
\begin{equation}
\label{eq:satt_acic}
\mathrm{SATT}
=\frac{1}{\sum_{t\in\mathcal{T}_{\mathrm{post}}} N_t}
\sum_{t\in\mathcal{T}_{\mathrm{post}}}\sum_{j:Z_j=1}
n_{jt}\,\{Y_{jt}(1)-Y_{jt}(0)\},
\end{equation}
where $\mathcal{T}_{\mathrm{post}}=\{3,4\}$ and $N_t = \sum_{j:Z_j=1} n_{jt}$ is the total number of treated patients at time $t$. Each causal inference algorithm on this dataset returns a point estimate $\hat{\tau}_r$ and a 90\% interval $[L_r, U_r]$ for replicate $r=1,\ldots,R$, and we evaluate
\begin{equation}
\label{eq:metric-acic}
\mathrm{RMSE}
= \sqrt{\frac{1}{R}\sum_{r=1}^{R}(\hat{\tau}_r - \tau_r^\star)^2},
\qquad
\mathrm{Coverage}_{0.90}
= \frac{1}{R}\sum_{r=1}^{R}\mathbf{1}\{\tau_r^\star \in [L_r, U_r]\},
\end{equation}
where $\tau_r^\star$ is the true replicate-level SATT. This RMSE is therefore computed over scalar SATT estimates across replicates and is not the unit-level $\sqrt{\mathrm{PEHE}}$. Our experiments use a randomly sampled subset of 150 out of the 300 released Track~2 replicates bundled with the original study, and we tested different train/test splits (30/120 for the main analysis and 75/75 for sensitivity analyses).

\textbf{ACIC 2016.} ACIC~2016 is a semi-synthetic cross-sectional benchmark built from a fixed real covariate matrix of 4{,}802 individuals with 58 pre-treatment covariates, while treatment assignment and potential outcomes are simulated under known data-generating mechanisms~\citep{dorie2019automated}. The released covariates include continuous, binary, count, and categorical variables. The released benchmark contains 200 semi-synthetic replicates (5 DGPs $\times$ 40 simulations). To keep evaluation budgets aligned across benchmarks, each run in our study first sampled a 100-replicate working subset from this pool and then split that subset as 20/80 for the prespecified main analysis or 50/50 for sensitivity analyses. For each replicate, we evaluate unit-level $\sqrt{\mathrm{PEHE}}$ using equation~\ref{eq:pehe-def} and absolute ATE error using equation~\ref{eq:ate-def}. 

\textbf{IHDP.} The Infant Health and Development Program benchmark draws covariates and treatment assignments from a randomized intervention study of low-birth-weight, premature infants, with simulated counterfactual outcomes~\citep{hill_bayesian_2011}. The benchmark version used in our study contains 25 pre-treatment covariates and 50 released replicates; each run uses a 25-replicate working subset, with either 12/13 or 5/20 train/validation splits.

\textbf{LaLonde.} The Lalonde dataset~\citep{lalonde1986evaluating} comes from the National Supported Work study, which investigated the effectiveness of a job training program on individuals' real earnings a few years after program completion. The version we used has eight pre-treatment covariates as well as simulated counterfactual outcomes mimicking the real data~\citep{dehejia1999causal}. We evaluated 50-replicate working subsets sampled from a pool of 100 replicates (10/40 for the main analysis and 25/25 for sensitivity analyses) using different algorithms' $\sqrt{\mathrm{PEHE}}$ and absolute ATE error. Since earnings are measured on a much larger dollar scale than in IHDP or ACIC~2016, the main text reports a normalized LaLonde variant in which both $\sqrt{\mathrm{PEHE}}$ and $|\text{ATE error}|$ are divided by the pooled standard deviation of the true ITEs across the evaluated replicates; the supplementary material reports the raw-scale sensitivity analysis.

\subsection*{\MethodName{} framework}

\MethodName{} builds on OpenEvolve~\citep{openevolve}, an implementation of evolutionary program synthesis that uses MAP-Elites~\citep{mouret2015illuminating} to preserve both high-performing and diverse solutions. The exact loop used in every run is summarized in Algorithm~\ref{alg:self-evolving-main}. Search-time scoring is always computed on the training replicates only; the final archived best program is then evaluated once on the untouched validation replicates, and only these held-out metrics appear in the main text.

\begin{algorithm}[t]
  \caption{\MethodName{} workflow for causal estimator discovery}
  \label{alg:self-evolving-main}
  \begin{algorithmic}[1]
    \REQUIRE Seed program $P_0$, evaluator $E$, benchmark replicates $\mathcal{D}$, configuration $C$
    \STATE Split $\mathcal{D}$ into training replicates $\mathcal{D}_{\mathrm{train}}$ and held-out validation replicates $\mathcal{D}_{\mathrm{val}}$.
    \STATE Initialize MAP-Elites archive $\mathcal{A}$ with $(P_0, E(P_0,\mathcal{D}_{\mathrm{train}}))$.
    \FOR{$i = 1, \dots, C.\mathrm{max\_iterations}$}
      \STATE Sample a parent program and inspiration set from $\mathcal{A}$.
      \STATE Compose a prompt containing code, scores, and recent differences; sample an LLM from the ensemble.
      \STATE Generate a candidate program $P'$.
      \STATE Run fail-fast evaluation on a small subset of $\mathcal{D}_{\mathrm{train}}$.
      \IF{$P'$ passes the threshold}
        \STATE Run full evaluation on all of $\mathcal{D}_{\mathrm{train}}$ and compute the dataset-specific composite score.
        \STATE Insert $(P', \mathrm{metrics}, S(P'))$ into $\mathcal{A}$ and update the current best program.
      \ENDIF
    \ENDFOR
    \STATE Evaluate the final best archived program once on $\mathcal{D}_{\mathrm{val}}$.
  \end{algorithmic}
\end{algorithm}

For search-time selection, each dataset uses a scalarized score that balances primary accuracy with a secondary objective:
\begin{equation}
S_{\text{ACIC2022}} = \frac{1}{1 + \log(1 + \mathrm{RMSE}) + \lambda \log\!\left(1 + \frac{|\mathrm{Coverage} - 0.9|}{0.9}\right)},
\end{equation}
\begin{equation}
S_{\text{ITE}} = \frac{1}{1 + \log(1 + \overline{\sqrt{\mathrm{PEHE}}}) + \lambda \log(1 + \overline{|\mathrm{ATE\ error}|})},
\end{equation}
where the overlines denote averages across the training replicates evaluated at that iteration. The score $S_{\text{ITE}}$ is the scalarized criterion used for ACIC~2016, IHDP, and LaLonde, whereas $S_{\text{ACIC2022}}$ is the customized metric used for the ACIC~2022 dataset given its different estimand and evaluation goals. These scalarizations are used only for archive selection; all main-text results are reported using the original held-out metrics: replicate-level SATT RMSE and coverage for ACIC~2022, and unit-level $\sqrt{\text{PEHE}}$ (i.e., the ITE RMSE) together with absolute ATE error for the cross-sectional benchmarks.

\subsection*{Proxy evaluation metrics}

While evaluation against ground-truth treatment effects in synthetic and semi-synthetic benchmarks remains the gold standard for comparing causal inference methods~\citep{cinelli2025challenges}, real-world applications do not provide such counterfactual truth. To enable \MethodName{} without access to hidden outcomes, we therefore optimize proxy objectives constructed only from observed data.

To do so, we draw on recent work on observed-data model-selection criteria and surrogate metrics for heterogeneous-effect estimation~\citep{mahajan2024empirical}, where researchers construct observed-data \emph{estimators} of treatment-effect error and then optimize those estimated errors. We detail the proxy objectives below. All proxy objectives use the same monotone normalization $\phi(x) = \log(1+x)/(1+\log(1+x))$ and are mapped back to a maximization score through $S = 1/(1 + \mathcal{L}_{\text{proxy}})$.

\textbf{ITE proxy objective (IHDP, ACIC~2016, LaLonde).} We construct cross-fitted doubly-robust pseudo-outcomes following \citet{kennedy2024semiparametric}. In the implemented evaluator, nuisance estimation uses two folds: within each fold, we estimate $\hat{\mu}_t(x)=\mathbb{E}[Y\mid X=x,T=t]$ for $t\in \{0,1\}$ using both Ridge and histogram gradient-boosted regressors, estimate the propensity model $\hat{e}(x)=P(T=1\mid X=x)$ with logistic regression clipped to $[0.05,0.95]$, and average the resulting doubly robust pseudo-outcomes across the two outcome-model families. The pseudo-outcome for unit $i$ is
\begin{equation}
\tilde{\tau}_i =
\hat{\mu}_1(x_i) - \hat{\mu}_0(x_i)
+ \frac{T_i(Y_i-\hat{\mu}_1(x_i))}{\hat{e}(x_i)}
- \frac{(1-T_i)(Y_i-\hat{\mu}_0(x_i))}{1-\hat{e}(x_i)}.
\end{equation}
The three proxy components are then: (i) DR pseudo-outcome fit for individual treatment effect predictions, measured as
\[
\widehat{\sqrt{\mathrm{PEHE}}}_{\text{proxy}} =
\sqrt{\frac{1}{N}\sum_i(\hat{\tau}_i - \tilde{\tau}_i)^2};
\]
(ii) proxy ATE error relative to the cross-fitted AIPW estimate $\hat{\tau}_{\text{AIPW}} = N^{-1}\sum_i \tilde{\tau}_i$; and (iii) normalized R-loss,
\begin{equation}
R_{\text{loss,norm}}
=
\frac{N^{-1}\sum_i\left(Y_i - \hat{m}(x_i) - (T_i-\hat{e}(x_i))\hat{\tau}_i\right)^2}
{\widehat{\mathrm{Var}}(Y) + 10^{-8}},
\end{equation}
where $\hat{m}(x)$ is a cross-fitted main-effect regression of $Y$ on $X$. The implemented objective is
\begin{equation}
\mathcal{L}_{\text{ITE-proxy}} =
w_{\text{PEHE}}\phi(\overline{\widehat{\sqrt{\text{PEHE}}}_{\text{proxy}}}) +
w_{\text{ATE}}\lambda_{\text{ATE}}\phi(\overline{|\hat{\tau} - \hat{\tau}_{\text{AIPW}}|}) +
w_R\phi(\overline{R_{\text{loss,norm}}}),
\end{equation}
with normalized weights $(w_{\text{PEHE}}, w_{\text{ATE}}, w_R) = (0.5, 0.3, 0.2)$. The final proxy scores are relatively stable to the weights specification. 

\textbf{ACIC~2022 proxy objective.} Because the ACIC~2022 task has a different estimand and evaluation target, we built its proxy objective around a doubly robust difference-in-differences (DR-DID) pseudo-target. The implemented proxy combined three terms: (i) absolute error of the point estimate relative to the DR-DID target; (ii) a coverage term based primarily on confidence-interval hit-rate against that target; and (iii) interval-width regularization to discourage degenerate overly wide intervals. When we had more than 20 replicates, the coverage term also included a secondary width-calibration penalty,
\[
g_{\text{width}} =
\frac{\left|\overline{W} - 2 z_{0.95}\,\widehat{\mathrm{sd}}(\hat{\tau}_{\text{DR-DID}})\right|}
{2 z_{0.95}\,\widehat{\mathrm{sd}}(\hat{\tau}_{\text{DR-DID}})},
\]
which compares the mean reported interval width with the nominal 90\% width implied by the cross-replicate spread of the DR-DID targets. This discourages confidence intervals that achieve nominal coverage only by becoming excessively wide. The implemented objective is
\begin{equation}
\mathcal{L}_{\text{ACIC-proxy}} =
w_{\tau}\phi(\overline{|\hat{\tau} - \hat{\tau}_{\text{DR-DID}}|}) +
w_{\text{cov}}\lambda_{\text{cov}}\widetilde{C} +
w_{\text{width}}\phi(\overline{W}),
\end{equation}
where $\widetilde{C} = \phi(|\hat{c} - 0.9|/0.9)$ if width calibration is unavailable and otherwise $\widetilde{C} = 0.7\,\phi(|\hat{c} - 0.9|/0.9) + 0.3\,\phi(g_{\text{width}})$. The normalized weights are $(w_{\tau}, w_{\text{cov}}, w_{\text{width}}) = (0.6, 0.25, 0.15)$.

\subsection*{Baseline Models}
We compared \MethodName{} against a fixed set of strong, widely used causal estimators, all evaluated on the same held-out benchmark replicates. For the cross-sectional benchmarks (IHDP, ACIC~2016, and LaLonde), the baseline suite comprised ordinary least squares adjustment, inverse propensity weighting (IPW; \citealp{rosenbaum1983central,austin2011introduction}), Causal Forest (\citealp{battocchi2019econml,wager2018estimation}), BART (\texttt{bartpy2}; \citealp{hill_bayesian_2011}), and an honest generalized random forest-style causal forest (\texttt{econml.grf}; \citealp{battocchi2019econml,athey2019generalized}). For ACIC~2022, where the target is a replicate-level panel SATT rather than unit-level ITEs, we used a backdoor-adjustment panel baseline and the ACIC~2022 human competition results (58 submissions) as the main external benchmark. For LaLonde, we additionally report the difference in means because it remains a standard design-based reference for this observational setting. Together, these baselines cover linear adjustment, weighting, Bayesian tree ensembles, and forest-based heterogeneous treatment effect estimators. In particular, they span the estimator families that modern causal-ML practitioners routinely consider on real observational or semi-synthetic datasets, and they overlap with the method classes implemented in popular AI-powered causal inference method-selection pipelines~\citep{verma2025causal}. We therefore treat them as strong representative package baselines, while noting that the paper does not attempt a separate benchmark-by-benchmark AutoML or hyperparameter-tuning sweep.

\subsection*{Experimental Design and Implementation Details}
To assess the effect of the main selection parameters in \MethodName{}, we implemented a factorial design experiment crossing four datasets, two primary LLM code generators from the GPT and Gemini families (GPT-5 and Gemini~3~Flash~Preview; \citealp{achiam2023gpt,team2024gemini}), three regularization strengths for combining two objectives ($\lambda \in \{0.2, 1.0, 5.0\}$), two held-out evaluation regimes (20:80 for the prespecified main analysis and 50:50 for sensitivity analyses), and two independent replicate seeds to assess variability across LLM generations, yielding 24 true-score final programs per dataset. All main-text figures and tables report the 20:80 slice of this design; the matched 50:50 reruns are summarized separately in Supplementary Section~\ref{sec:supp-heldout-sensitivity}. As recommended by the OpenEvolve pipeline, each primary LLM was paired with Claude~Sonnet~4 as a secondary model with 20\% weight in the ensemble configuration. We set the maximum number of evolution iterations to 100 for all experiments. Within a given dataset and split, we used identical training/validation partitions across model and regularization conditions. All search-time scoring used only the sampled training replicates. This replicate-split design partially mitigates benchmark-contamination concerns: even if a frontier model encodes general priors about IHDP or LaLonde, no run is rewarded for repeating a single memorized replicate, because the same replicate never appears in both search and final evaluation and all gains are judged relative to the zero-shot seed on held-out replicates. All LLM calls used the OpenRouter API with temperature 0.7, top-$p$ of 0.95, and a maximum of 8{,}192 output tokens.

We used the default MAP-Elites search parameters in OpenEvolve: population size 60, 4 islands, archive size 25, elite selection ratio 0.3, and exploitation ratio 0.7. The prompt sampler supplied the current parent together with the top~3 archived programs and 2 additional diverse programs as context. To encourage exploration of causal methods from different methodological families, we required candidate mutations to be generated as full program rewrites rather than patch-style diffs.

\textbf{Evaluation pipeline.} Candidate programs underwent a two-stage cascade evaluation to ensure efficiency of the pipeline. Stage~1 executed each program on approximately 10\% of the training replicates with a single worker; programs scoring below 0.001 were discarded to filter out candidates that errored on the inputs or did not return valid estimates. Stage~2 reevaluated the surviving programs on all training replicates, again using a single-worker subprocess evaluator during search. Each evaluation was subject to a benchmark-specific timeout of 1,800 seconds enforced via process-level termination; timed-out programs received a combined score of 0.0. This implicitly penalized programs that were too slow on these datasets to be practically useful. Programs returning non-finite values, mismatched output lengths, or runtime exceptions likewise received penalty scores.

\textbf{Prompts and allowed libraries.} Each dataset's evolution prompt specified the data schema, evaluation metric formula, and required \texttt{estimate(...)} function signature. Prompts instructed the LLM to produce pure, deterministic Python functions using only \texttt{pandas}, \texttt{numpy}, \texttt{scikit-learn}, and \texttt{statsmodels}, and to modify only the evolvable code block. To improve the robustness of LLM-generated code, the runtime pre-injected common imports and utility classes (for example, \texttt{numpy}, \texttt{pandas}, \texttt{StandardScaler}, \texttt{KFold}, \texttt{LogisticRegression}, \texttt{Ridge}, and \texttt{GradientBoostingRegressor}). Complete prompt templates are reproduced in Supplementary Section~\ref{sec:supp-evolution-prompts} and in the code repository.

\subsection*{Analysis of the generated programs}
To characterize the evolved programs beyond evaluation metrics, we assembled one final code snippet per program from all baseline, true-evolved, and proxy-evolved runs across the four benchmarks, comprising 28 baseline programs, 96 true-evolved programs, and 48 proxy-evolved programs (20:80 split only). We first compared code character counts and non-empty line counts.

As a complementary proxy for conceptual complexity, we translated each code snippet with three different LLMs (GPT-5, Gemini~2.5~Pro, and Claude~Sonnet~4.5) into a methods paragraph for a scientific manuscript. All three models received the same instruction: describe only what the code actually implements, do not speculate or discuss absent steps, and return plain text only. For each translated paragraph, we computed word, character, and sentence counts and quantified the association between source-code size and translated-paragraph length.

Finally, we analyzed the structural diversity of the evolved programs. To assess whether evolved programs rediscover known estimators, we built a commit-pinned reference library of seven official wrappers---one per estimator family (T-learner, X-learner, DR-learner, IPW, OLS, AIPW, and causal forest)---adapted from maintainer examples in the \texttt{EconML}, \texttt{zEpid}, and \texttt{statsmodels} repositories and normalized to a shared \texttt{estimate(df)} interface. For each evolved program, we computed the nearest code-repository neighbor based on cosine similarity using TF-IDF (with \texttt{text-embedding-3-large} used for sensitivity analysis). We also computed pairwise TF-IDF cosine similarities (and likewise with \texttt{text-embedding-3-large}) among all final programs to summarize within- and between-dataset code diversity. 

To classify each program's algorithm family and extract novel components, we ran an LLM-as-judge pipeline with three high-performing LLMs (GPT-5, Gemini~2.5~Pro, and Claude~Sonnet~4.5), aggregated their outputs, and used majority vote as the final label. These results were reviewed by two graduate students with causal-inference expertise before summarization. Supplementary Section~\ref{sec:supp-program-analysis} summarizes the exact program pool, the translation and judge prompts, and the alternate-embedding sensitivity analysis.

The supplementary material is structured to mirror the main narrative. Supplementary Section~\ref{sec:supp-search-setup} documents the exact prompts, search loop, full hyperparameter grid, score definitions, and implementation controls (\ref{sec:supp-evolution-prompts}--\ref{sec:supp-implementation-details}); Supplementary Section~\ref{sec:supp-benchmark-summaries} collects the full zero-shot tables, program-analysis prompts and source manifests, code-similarity summaries, and complete complexity figures (\ref{sec:supp-zero-shot-tables}--\ref{sec:supp-program-lengths}). Supplementary Section~\ref{sec:supp-sensitivity-proxy} covers split sensitivity, $\lambda$ sensitivity, proxy construction, and proxy validation (\ref{sec:supp-heldout-sensitivity}--\ref{sec:supp-proxy-validation}), Supplementary Section~\ref{sec:supp-trajectories} reports representative dataset-level trajectories (\ref{sec:supp-acic2022-trajectory}--\ref{sec:supp-lalonde-trajectory}), and Supplementary Section~\ref{sec:supp-code-edits} reproduces representative code mutations (\ref{sec:supp-acic2022-edit-ridge}--\ref{sec:supp-lalonde-edit}).

\begin{summarybox}{Program-analysis workflow}
\textbf{Complexity measures.} We quantified source-code size directly and translated every program into manuscript-style methods prose with three frontier LLMs to obtain a second, text-based complexity measure.\par\smallskip
\textbf{Reference matching.} We compared each evolved program against seven commit-pinned official reference wrappers from \texttt{EconML}, \texttt{zEpid}, and \texttt{statsmodels} using TF-IDF and \texttt{text-embedding-3-large} cosine similarity, and contextualized those values against similarity among the references themselves.\par\smallskip
\textbf{Judge ensemble.} A three-model judge panel produced the method taxonomy, nearest-known-method labels, and extracted novel components used in the program-structure analysis.
\end{summarybox}
}

\ifdefined\ArxivMethodsFirst
  \MethodsSectionContent
  \ResultsSectionContent
  \DiscussionSectionContent
\else
  \ResultsSectionContent
  \DiscussionSectionContent
  \MethodsSectionContent
\fi

\section*{Data availability}
The benchmark datasets used in this study are publicly available from their original sources: ACIC~2022 (\url{https://acic2022.mathematica.org/data-challenge/}), ACIC~2016 (\url{https://github.com/vdorie/aciccomp}), the IHDP benchmark replications (\url{https://github.com/AMLab-Amsterdam/CEVAE}), and the Dehejia--Wahba / LaLonde data distributed through CRAN packages such as \texttt{Matching} (\url{https://search.r-project.org/CRAN/refmans/Matching/html/lalonde.html}). Exact download instructions, preprocessing steps, and benchmark-specific wrappers are documented in the project repository. An interactive exploration of the benchmark results and evolved programs is available at \url{https://yiqunchen.github.io/causal-agent/}.

\section*{Code availability}
Code for \MethodName{}, including prompts, evaluation scripts, figure generation, and analysis workflows, is available at \url{https://github.com/yiqunchen/causal-agent}. The repository also documents dataset access, preprocessing, per-run configurations, and the final evolved programs used in this manuscript.

\section*{Acknowledgements}
We would like to thank the Causal Inference Working Group in the Department of Biostatistics at Johns Hopkins University and Alex Luedtke for helpful feedback. This work was partially supported by a Data Science and AI Faculty Innovation Fund in the Department of Biostatistics at the Johns Hopkins Bloomberg School of Public Health.


\bibliography{references}

\begin{thebibliography}{42}
\providecommand{\natexlab}[1]{#1}
\providecommand{\url}[1]{\texttt{#1}}
\expandafter\ifx\csname urlstyle\endcsname\relax
  \providecommand{\doi}[1]{doi: #1}\else
  \providecommand{\doi}{doi: \begingroup \urlstyle{rm}\Url}\fi

\bibitem[Acharya et~al.(2025)Acharya, Zhang, Kim, Haghighat, Sun, Shrestha,
  Mordig, Danisman, Jose, Qi, Cobben, Schölkopf, Sachan, and
  Jin]{acharya_causcibench_2025}
Sawal Acharya, Terry~Jingchen Zhang, Andrew Kim, Anahita Haghighat, Xianlin
  Sun, Rahul~Babu Shrestha, Maximilian Mordig, Furkan Danisman, Clijo Jose,
  Yahang Qi, Pepijn Cobben, Bernhard Schölkopf, Mrinmaya Sachan, and Zhijing
  Jin.
\newblock {CauSciBench}: {Assessing} {LLM} {Causal} {Reasoning} for
  {Scientific} {Research}.
\newblock \emph{Advances in Neural Information Processing Systems}, 38, October
  2025.
\newblock URL \url{https://openreview.net/forum?id=EO8mTLqDuT}.

\bibitem[Achiam et~al.(2023)Achiam, Adler, Agarwal, Ahmad, Akkaya, Aleman,
  Almeida, Altenschmidt, Altman, Anadkat, et~al.]{achiam2023gpt}
Josh Achiam, Steven Adler, Sandhini Agarwal, Lama Ahmad, Ilge Akkaya,
  Florencia~Leoni Aleman, Diogo Almeida, Janko Altenschmidt, Sam Altman,
  Shyamal Anadkat, et~al.
\newblock {GPT-4 technical report}.
\newblock \emph{arXiv preprint arXiv:2303.08774}, 2023.

\bibitem[Athey et~al.(2019)Athey, Tibshirani, and Wager]{athey2019generalized}
Susan Athey, Julie Tibshirani, and Stefan Wager.
\newblock Generalized random forests.
\newblock \emph{The Annals of Statistics}, 47\penalty0 (2):\penalty0
  1148--1178, 2019.
\newblock \doi{10.1214/18-AOS1709}.

\bibitem[Austin(2011)]{austin2011introduction}
Peter~C Austin.
\newblock An introduction to propensity score methods for reducing the effects
  of confounding in observational studies.
\newblock \emph{Multivariate Behavioral Research}, 46\penalty0 (3):\penalty0
  399--424, 2011.

\bibitem[Bailey et~al.(2024)Bailey, Jung, Beltz, Eronen, Gische, Hamaker,
  Kording, Lebel, Lindquist, Moeller, et~al.]{bailey2024causal}
Drew~H Bailey, Alexander~J Jung, Adriene~M Beltz, Markus~I Eronen, Christian
  Gische, Ellen~L Hamaker, Konrad~P Kording, Catherine Lebel, Martin~A
  Lindquist, Julia Moeller, et~al.
\newblock Causal inference on human behaviour.
\newblock \emph{Nature Human Behaviour}, 8\penalty0 (8):\penalty0 1448--1459,
  2024.

\bibitem[Ban et~al.(2025)Ban, Chen, Lyu, Wang, Zhu, Tu, and
  Chen]{ban2025integrating}
Taiyu Ban, Lyuzhou Chen, Derui Lyu, Xiangyu Wang, Qinrui Zhu, Qiang Tu, and
  Huanhuan Chen.
\newblock Integrating large language model for improved causal discovery.
\newblock \emph{IEEE Transactions on Artificial Intelligence}, 2025.

\bibitem[Bang and Robins(2005)]{bang2005doubly}
Heejung Bang and James~M Robins.
\newblock Doubly robust estimation in missing data and causal inference models.
\newblock \emph{Biometrics}, 61\penalty0 (4):\penalty0 962--973, 2005.

\bibitem[Battocchi et~al.(2019)Battocchi, Dillon, Hei, Lewis, Oka, Oprescu, and
  Syrgkanis]{battocchi2019econml}
Keith Battocchi, Eleanor Dillon, Maggie Hei, Greg Lewis, Paul Oka, Miruna
  Oprescu, and Vasilis Syrgkanis.
\newblock {EconML}: {A Python Package for ML-Based Heterogeneous Treatment
  Effects Estimation}.
\newblock GitHub repository, 2019.
\newblock URL \url{https://github.com/py-why/EconML}.

\bibitem[Bazgir et~al.(2025)Bazgir, Habibdoust, Zhang, and
  Song]{bazgir_causal_2025}
Adib Bazgir, Amir Habibdoust, Yuwen Zhang, and Xing Song.
\newblock Causal {MAS}: {A} {Survey} of {Large} {Language} {Model}
  {Architectures} for {Discovery} and {Effect} {Estimation}, August 2025.
\newblock URL \url{http://arxiv.org/abs/2509.00987}.
\newblock arXiv:2509.00987 [cs].

\bibitem[Chernozhukov et~al.(2018)Chernozhukov, Chetverikov, Demirer, Duflo,
  Hansen, Newey, and Robins]{chernozhukov2018double}
Victor Chernozhukov, Denis Chetverikov, Mert Demirer, Esther Duflo, Christian
  Hansen, Whitney Newey, and James Robins.
\newblock Double/debiased machine learning for treatment and structural
  parameters, 2018.

\bibitem[Cinelli et~al.(2025)Cinelli, Feller, Imbens, Kennedy, Magliacane, and
  Zubizarreta]{cinelli2025challenges}
Carlos Cinelli, Avi Feller, Guido Imbens, Edward Kennedy, Sara Magliacane, and
  Jose Zubizarreta.
\newblock Challenges in statistics: A dozen challenges in causality and causal
  inference.
\newblock \emph{arXiv preprint arXiv:2508.17099}, 2025.

\bibitem[Dehejia and Wahba(1999)]{dehejia1999causal}
Rajeev~H Dehejia and Sadek Wahba.
\newblock Causal effects in nonexperimental studies: Reevaluating the
  evaluation of training programs.
\newblock \emph{Journal of the American Statistical Association}, 94\penalty0
  (448):\penalty0 1053--1062, 1999.

\bibitem[Dorie et~al.(2019)Dorie, Hill, Shalit, Scott, and
  Cervone]{dorie2019automated}
Vincent Dorie, Jennifer Hill, Uri Shalit, Marc Scott, and Dan Cervone.
\newblock Automated versus do-it-yourself methods for causal inference: Lessons
  learned from a data analysis competition.
\newblock \emph{Statistical Science}, 34\penalty0 (1):\penalty0 43--68, 2019.

\bibitem[Feuerriegel et~al.(2024)Feuerriegel, Frauen, Melnychuk, Schweisthal,
  Hess, Curth, Bauer, Kilbertus, Kohane, and van~der
  Schaar]{feuerriegel2024causal}
Stefan Feuerriegel, Dennis Frauen, Valentyn Melnychuk, Jonas Schweisthal,
  Konstantin Hess, Alicia Curth, Stefan Bauer, Niki Kilbertus, Isaac~S Kohane,
  and Mihaela van~der Schaar.
\newblock Causal machine learning for predicting treatment outcomes.
\newblock \emph{Nature Medicine}, 30\penalty0 (4):\penalty0 958--968, 2024.

\bibitem[Gao et~al.(2024)Gao, Fang, Huang, Giunchiglia, Noori, Schwarz,
  Ektefaie, Kondic, and Zitnik]{gao2024empowering}
Shanghua Gao, Ada Fang, Yepeng Huang, Valentina Giunchiglia, Ayush Noori,
  Jonathan~Richard Schwarz, Yasha Ektefaie, Jovana Kondic, and Marinka Zitnik.
\newblock Empowering biomedical discovery with ai agents.
\newblock \emph{Cell}, 187\penalty0 (22):\penalty0 6125--6151, 2024.

\bibitem[Gottweis et~al.(2025)Gottweis, Weng, Daryin, Tu, Palepu, Sirkovic, and
  Natarajan]{gottweis2025towards}
J~Gottweis, WH~Weng, A~Daryin, T~Tu, A~Palepu, P~Sirkovic, and V~Natarajan.
\newblock Towards an ai co-scientist: A multi-agent system for scientific
  discovery.
\newblock \emph{arXiv preprint arXiv:2502.18864}, page~3, 2025.

\bibitem[Hern{\'a}n and Robins(2010)]{hernan2010causal}
Miguel~A Hern{\'a}n and James~M Robins.
\newblock Causal inference, 2010.

\bibitem[Hill(2011)]{hill_bayesian_2011}
Jennifer~L. Hill.
\newblock Bayesian {Nonparametric} {Modeling} for {Causal} {Inference}.
\newblock \emph{Journal of Computational and Graphical Statistics}, 20\penalty0
  (1):\penalty0 217--240, January 2011.
\newblock ISSN 1061-8600.
\newblock \doi{10.1198/jcgs.2010.08162}.
\newblock URL \url{https://doi.org/10.1198/jcgs.2010.08162}.
\newblock Publisher: ASA Website \_eprint:
  https://doi.org/10.1198/jcgs.2010.08162.

\bibitem[Imbens and Rubin(2015)]{imbens2015causal}
Guido~W Imbens and Donald~B Rubin.
\newblock \emph{Causal inference in statistics, social, and biomedical
  sciences}.
\newblock Cambridge university press, 2015.

\bibitem[Karpathy(2026)]{karpathy_autoresearch_2026}
Andrej Karpathy.
\newblock autoresearch, 2026.
\newblock URL \url{https://github.com/karpathy/autoresearch}.
\newblock GitHub repository, commit COMMIT, accessed 2026-04-05.

\bibitem[Kennedy(2023)]{kennedy2023towards}
Edward~H Kennedy.
\newblock Towards optimal doubly robust estimation of heterogeneous causal
  effects.
\newblock \emph{Electronic Journal of Statistics}, 17\penalty0 (2):\penalty0
  3008--3049, 2023.

\bibitem[Kennedy(2024)]{kennedy2024semiparametric}
Edward~H Kennedy.
\newblock Semiparametric doubly robust targeted double machine learning: a
  review.
\newblock \emph{Handbook of Statistical Methods for Precision Medicine}, pages
  207--236, 2024.

\bibitem[K{\"u}nzel et~al.(2019)K{\"u}nzel, Sekhon, Bickel, and
  Yu]{kunzel2019metalearners}
S{\"o}ren~R K{\"u}nzel, Jasjeet~S Sekhon, Peter~J Bickel, and Bin Yu.
\newblock Metalearners for estimating heterogeneous treatment effects using
  machine learning.
\newblock \emph{Proceedings of the National Academy of Sciences}, 116\penalty0
  (10):\penalty0 4156--4165, 2019.

\bibitem[LaLonde(1986)]{lalonde1986evaluating}
Robert~J LaLonde.
\newblock Evaluating the econometric evaluations of training programs with
  experimental data.
\newblock \emph{The American Economic Review}, 76\penalty0 (4):\penalty0
  604--620, 1986.

\bibitem[Mahajan et~al.(2024)Mahajan, Mitliagkas, Neal, and
  Syrgkanis]{mahajan2024empirical}
Divyat Mahajan, Ioannis Mitliagkas, Brady Neal, and Vasilis Syrgkanis.
\newblock Empirical analysis of model selection for heterogeneous causal effect
  estimation.
\newblock In \emph{The Twelfth International Conference on Learning
  Representations}, 2024.
\newblock URL \url{https://openreview.net/forum?id=yuy6cGt3KL}.

\bibitem[Miettinen(1999)]{miettinen1999nonlinear}
Kaisa Miettinen.
\newblock \emph{Nonlinear multiobjective optimization}, volume~12.
\newblock Springer Science \& Business Media, 1999.

\bibitem[Miliani et~al.(2025)Miliani, Auriemma, Bondielli, Chersoni, Passaro,
  Sucameli, and Lenci]{miliani2025explica}
Martina Miliani, Serena Auriemma, Alessandro Bondielli, Emmanuele Chersoni,
  Lucia Passaro, Irene Sucameli, and Alessandro Lenci.
\newblock Explica: Evaluating explicit causal reasoning in large language
  models.
\newblock In \emph{Findings of the Association for Computational Linguistics:
  ACL 2025}, pages 17335--17355, 2025.

\bibitem[Mouret and Clune(2015)]{mouret2015illuminating}
Jean-Baptiste Mouret and Jeff Clune.
\newblock Illuminating search spaces by mapping elites.
\newblock \emph{arXiv preprint arXiv:1504.04909}, 2015.

\bibitem[Novikov et~al.(2025)Novikov, V{\~u}, Eisenberger, Dupont, Huang,
  Wagner, Shirobokov, Kozlovskii, Ruiz, Mehrabian,
  et~al.]{novikov2025alphaevolve}
Alexander Novikov, Ng{\^a}n V{\~u}, Marvin Eisenberger, Emilien Dupont, Po-Sen
  Huang, Adam~Zsolt Wagner, Sergey Shirobokov, Borislav Kozlovskii,
  Francisco~JR Ruiz, Abbas Mehrabian, et~al.
\newblock Alphaevolve: A coding agent for scientific and algorithmic discovery.
\newblock \emph{arXiv preprint arXiv:2506.13131}, 2025.

\bibitem[Pearl(2009)]{pearl2009causal}
Judea Pearl.
\newblock Causal inference in statistics: An overview.
\newblock \emph{Statistics Surveys}, 3:\penalty0 96--146, 2009.

\bibitem[Rosenbaum and Rubin(1983)]{rosenbaum1983central}
Paul~R Rosenbaum and Donald~B Rubin.
\newblock The central role of the propensity score in observational studies for
  causal effects.
\newblock \emph{Biometrika}, 70\penalty0 (1):\penalty0 41--55, 1983.

\bibitem[Sharma(2025)]{openevolve}
Asankhaya Sharma.
\newblock Openevolve: an open-source evolutionary coding agent.
\newblock \url{https://github.com/codelion/openevolve}, 2025.

\bibitem[Shen et~al.(2025)Shen, Chen, Luo, Xu, Chen, and Ni]{shen2025exploring}
ChengAo Shen, Zhengzhang Chen, Dongsheng Luo, Dongkuan Xu, Haifeng Chen, and
  Jingchao Ni.
\newblock Exploring multi-modal data with tool-augmented llm agents for precise
  causal discovery.
\newblock In \emph{Findings of the Association for Computational Linguistics:
  ACL 2025}, pages 636--660, 2025.

\bibitem[Stuart(2010)]{stuart2010matching}
Elizabeth~A Stuart.
\newblock Matching methods for causal inference: A review and a look forward.
\newblock \emph{Statistical Science}, 25\penalty0 (1):\penalty0 1, 2010.

\bibitem[Swanson et~al.(2025)Swanson, Wu, Bulaong, Pak, and
  Zou]{swanson2025virtual}
Kyle Swanson, Wesley Wu, Nash~L Bulaong, John~E Pak, and James Zou.
\newblock The virtual lab of ai agents designs new sars-cov-2 nanobodies.
\newblock \emph{Nature}, pages 1--3, 2025.

\bibitem[Team et~al.(2024)Team, Georgiev, Lei, Burnell, Bai, Gulati, Tanzer,
  Vincent, Pan, Wang, et~al.]{team2024gemini}
Gemini Team, Petko Georgiev, Ving~Ian Lei, Ryan Burnell, Libin Bai, Anmol
  Gulati, Garrett Tanzer, Damien Vincent, Zhufeng Pan, Shibo Wang, et~al.
\newblock Gemini 1.5: Unlocking multimodal understanding across millions of
  tokens of context.
\newblock \emph{arXiv preprint arXiv:2403.05530}, 2024.

\bibitem[Thal and Finucane(2023)]{thal_causal_2023}
Dan~R.C. Thal and Mariel~M. Finucane.
\newblock Causal {Methods} {Madness}: {Lessons} {Learned} from the 2022 {ACIC}
  {Competition} to {Estimate} {Health} {Policy} {Impacts}.
\newblock \emph{Observational Studies}, 9\penalty0 (3):\penalty0 3--27, 2023.
\newblock ISSN 2767-3324.
\newblock URL \url{https://muse.jhu.edu/pub/56/article/895650}.
\newblock Publisher: University of Pennsylvania Press.

\bibitem[Verma et~al.(2025)Verma, Acharya, Bhardwaj, Simko, Yang, Haghighat,
  Janzing, Sachan, Sch{\"o}lkopf, and Jin]{verma2025causal}
Vishal Verma, Sawal Acharya, Devansh Bhardwaj, Samuel Simko, Yongjin Yang,
  Anahita Haghighat, Dominik Janzing, Mrinmaya Sachan, Bernhard Sch{\"o}lkopf,
  and Zhijing Jin.
\newblock Causal {AI} scientist: Facilitating causal data science with large
  language models.
\newblock In \emph{NeurIPS 2025 Workshop on CauScien: Uncovering Causality in
  Science}, 2025.
\newblock URL \url{https://openreview.net/forum?id=EDWTHMVOCj}.

\bibitem[Wager and Athey(2018)]{wager2018estimation}
Stefan Wager and Susan Athey.
\newblock Estimation and inference of heterogeneous treatment effects using
  random forests.
\newblock \emph{Journal of the American Statistical Association}, 113\penalty0
  (523):\penalty0 1228--1242, 2018.

\bibitem[Wang(2024)]{wang2024causalbench}
Zeyu Wang.
\newblock Causalbench: A comprehensive benchmark for evaluating causal
  reasoning capabilities of large language models.
\newblock In \emph{Proceedings of the 10th SIGHAN Workshop on Chinese Language
  Processing (SIGHAN-10)}, pages 143--151, 2024.

\bibitem[Yamada et~al.(2025)Yamada, Lange, Lu, Hu, Lu, Foerster, Clune, and
  Ha]{yamada2025ai}
Yutaro Yamada, Robert~Tjarko Lange, Cong Lu, Shengran Hu, Chris Lu, Jakob
  Foerster, Jeff Clune, and David Ha.
\newblock The {AI} scientist-v2: Workshop-level automated scientific discovery
  via agentic tree search.
\newblock \emph{arXiv preprint arXiv:2504.08066}, 2025.

\bibitem[Zheng et~al.(2023)Zheng, Chiang, Sheng, Zhuang, Wu, Zhuang, Lin, Li,
  Li, Xing, et~al.]{zheng2023judging}
Lianmin Zheng, Wei-Lin Chiang, Ying Sheng, Siyuan Zhuang, Zhanghao Wu, Yonghao
  Zhuang, Zi~Lin, Zhuohan Li, Dacheng Li, Eric Xing, et~al.
\newblock {Judging LLM-as-a-judge with MT-bench and Chatbot Arena}.
\newblock \emph{Advances in Neural Information Processing Systems},
  36:\penalty0 46595--46623, 2023.

\end{thebibliography}


\clearpage
\section*{Supplementary Material}

This supplementary section is organized to parallel the main text. We first document the exact prompts, search workflow, and scoring rules that define the estimator-discovery problem. We then collect the additional benchmark summaries and sensitivity analyses referenced from the Results section, before turning to dataset-specific evolutionary trajectories and representative code edits.

\setcounter{table}{0}
\renewcommand{\thetable}{\arabic{table}}
\renewcommand{\tablename}{Supplementary Table}
\renewcommand{\theHtable}{supp.table.\arabic{table}}
\setcounter{figure}{0}
\renewcommand{\thefigure}{\arabic{figure}}
\renewcommand{\figurename}{Supplementary Fig.}
\renewcommand{\theHfigure}{supp.fig.\arabic{figure}}
\setcounter{algorithm}{0}
\floatname{algorithm}{Supplementary Algorithm}
\renewcommand{\theHalgorithm}{supp.alg.\arabic{algorithm}}

\subsection*{S1. Search setup and prompts}
\manualsectionlabel{sec:supp-search-setup}{S1}

This section defines the estimator-search problem solved by \MethodName{}. We begin with the exact prompt templates supplied to the LLMs, then summarize the common workflow, the fixed configuration choices, the benchmark-specific scoring rules, and the implementation details needed to reproduce the experiment.

\subsubsection*{S1.1. Evolution prompts}
\manualsectionlabel{sec:supp-evolution-prompts}{S1.1}

The four prompt templates below were used to initialize the evolutionary search across benchmarks. Each prompt fixes the data schema, return signature, evaluation metric, allowed libraries, and mutation rules; a short data preview was appended dynamically at runtime.

\clearpage
\begin{promptbox}{ACIC 2022 Evolution Prompt}
\begin{lstlisting}[breaklines=true, basicstyle=\scriptsize\ttfamily, columns=fullflexible, numbers=none, frame=none]
You are an expert causal inference statistician and Python programmer.

Background:
- We have multisite randomized trial data in long format (practice-year level).
- Treatment is assigned at the practice level, randomized, and 'post' marks the
  period after treatment rollout.

Data columns:
    - 'id.practice' (int): site identifier
    - 'year' (int): panel time index
    - 'Y' (float): average monthly spending per patient
    - 'Z' (0/1): treatment indicator
    - 'post' (0/1): post-treatment period indicator
    - 'n.patients' (int): patient count (weights)
    - Practice-level covariates: X1..X9
    - Aggregated patient covariates: V1_avg..V5_C_avg

Task: Estimate the Overall SATT for treated practices in the post-treatment
period, weighted by n.patients.

OpenEvolve Objective:
- Score = 1/(1 + log(1+RMSE) + lambda * log(1+coverage_gap)).
- Minimise RMSE while achieving close to 90% coverage.

Function signature:
    def estimate(df) -> (tau_hat, lb, ub, method)

Constraints: Pure, deterministic. Use only pandas, numpy, scikit-learn,
statsmodels. Robust to NaNs and edge cases. Modify only the EVOLVE-BLOCK.

[Data preview appended at runtime]
\end{lstlisting}
\end{promptbox}

\clearpage
\begin{promptbox}{IHDP Evolution Prompt}
\begin{lstlisting}[breaklines=true, basicstyle=\scriptsize\ttfamily, columns=fullflexible, numbers=none, frame=none]
You are an expert causal inference researcher and Python programmer working on
the IHDP semi-synthetic benchmark.

Dataset schema: `treatment` (binary), `y_factual` (outcome), `x1`..`x25`
(covariates). Ground-truth columns removed before execution.

Task: Implement a deterministic estimator returning (ite_hat, ate_hat,
method_name).

Evaluation: Score = 1/(1 + log(1+mean_sqrt_PEHE) + lambda*log(1+mean_ATE_err)).
50 replicates; configurable train/test split.

Objective: Minimise score. Explore diverse estimator families (T-/X-/DR-learners,
doubly-robust, etc.). Prioritise stability.

Requirements: Pure function. pandas, numpy, scikit-learn, statsmodels only.
Modify only the EVOLVE-BLOCK.

[Data preview appended at runtime]
\end{lstlisting}
\end{promptbox}

\clearpage
\begin{promptbox}{ACIC 2016 Evolution Prompt}
\begin{lstlisting}[breaklines=true, basicstyle=\scriptsize\ttfamily, columns=fullflexible, numbers=none, frame=none]
You are an expert causal inference researcher and Python programmer working on
the ACIC 2016 semi-synthetic benchmark.

Dataset schema: `treatment` (binary), `y_factual` (outcome), `x_1`..`x_58`
(covariates, mix of continuous and categorical). Ground-truth columns removed.

Task: Implement a deterministic estimator returning (ite_hat, ate_hat,
method_name).

Evaluation: source pool of 200 replicates (5 DGPs x 40 simulations);
main study uses a 100-replicate working subset per run.
Score = 1/(1 + log(1+mean_sqrt_PEHE) + lambda*log(1+mean_ATE_err)).

Objective: Minimise score. Explore diverse estimator families. Handle non-numeric
covariates robustly.

Requirements: Pure function. pandas, numpy, scikit-learn, statsmodels only.
Modify only the EVOLVE-BLOCK.

[Data preview appended at runtime]
\end{lstlisting}
\end{promptbox}

\clearpage
\begin{promptbox}{LaLonde Evolution Prompt}
\begin{lstlisting}[breaklines=true, basicstyle=\scriptsize\ttfamily, columns=fullflexible, numbers=none, frame=none]
You are an expert causal inference researcher and Python programmer working on
the LaLonde observational dataset (Dehejia-Wahba / PSID controls).

Dataset schema: `treatment` (binary), `y_factual` (re78 earnings, $0-$107K),
covariates: age, education, black, hispanic, married, nodegree, re74, re75.

Domain guidance:
- Observational study with strong selection bias between groups.
- Prior earnings (re74, re75) are strongest predictors.
- Consider propensity-score trimming and overlap enforcement.
- Do not assume ATE is positive.

Task: Implement a deterministic estimator returning (ite_hat, ate_hat,
method_name).

Evaluation: 100 replicates.
Score = 1/(1 + log(1+mean_sqrt_PEHE) + lambda*log(1+mean_ATE_err)).

Requirements: Pure function. pandas, numpy, scikit-learn, statsmodels only.
Modify only the EVOLVE-BLOCK.

[Data preview appended at runtime]
\end{lstlisting}
\end{promptbox}

\clearpage
\subsubsection*{S1.2. Algorithmic workflow}
\manualsectionlabel{sec:supp-algorithmic-workflow}{S1.2}

Algorithm~\ref{alg:self-evolving} summarizes the common search loop used across all 96 runs in the factorial experiment.

\begin{algorithm}[htbp]
  \caption{\MethodName{} workflow for causal estimator discovery}
  \label{alg:self-evolving}
  \begin{algorithmic}[1]
    \REQUIRE Baseline program $P_0$, evaluator $E$, configuration $C$, benchmark replicates $\mathcal{D}$
    \ENSURE Best program $P^\star$, evolution traces
    \STATE Seed all RNGs with global seed; split $\mathcal{D}$ into training $\mathcal{D}_{\text{train}}$ and held-out $\mathcal{D}_{\text{val}}$.
    \STATE Initialize the MAP-Elites archive $\mathcal{A}$ with $(P_0, E(P_0, \mathcal{D}_{\text{train}}))$.
    \STATE Build LLM ensemble: $\{(w_1, \text{primary model}), (w_2, \text{secondary model})\}$.
    \FOR{$i = 1 \dots C.\text{max\_iterations}$}
      \STATE $(P_{\text{parent}}, \mathcal{I}) \leftarrow$ sample parent and inspiration set from $\mathcal{A}$.
      \STATE $q \leftarrow$ \textsc{ComposePrompt}$(P_{\text{parent}}, \mathcal{I}, \text{metrics}, \text{top-}k\text{ programs})$.
      \STATE Sample LLM $\ell$ from ensemble with probabilities $(w_1, w_2)$.
      \STATE $P' \leftarrow \ell(q)$ \COMMENT{Generate candidate program}
      \STATE \textbf{Stage 1 (fail-fast):} Run $E(P', \mathcal{D}_{\text{train}}^{(1)})$ on ${\sim}10$\% of training replicates (timeout 180\,s).
      \IF{score $< \theta_1$}
        \STATE Discard $P'$; \textbf{continue}.
      \ENDIF
      \STATE \textbf{Stage 2 (full):} Evaluate $E(P', \mathcal{D}_{\text{train}})$ on all training replicates with 1 parallel worker (timeout 1800\,s).
      \STATE Compute composite score $S(P')$ using dataset-specific scoring function.
      \STATE Insert $(P', S, \text{metrics})$ into $\mathcal{A}$; update MAP-Elites bins and $P^\star$.
    \ENDFOR
    \STATE \textbf{Final evaluation:} Run $E(P^\star, \mathcal{D}_{\text{val}})$ on held-out replicates with 4 parallel workers (timeout 1800\,s).
    \RETURN $P^\star$, evolution traces, evaluation logs.
  \end{algorithmic}
\end{algorithm}
\FloatBarrier

\subsubsection*{S1.3. Full hyperparameter configuration}
\manualsectionlabel{sec:supp-full-config}{S1.3}

Supplementary Table~\ref{tab:supp-config} lists the fixed settings and factorial factors used throughout the full benchmark sweep.

\begin{table}[htbp]
\centering
\caption{\textbf{Complete hyperparameter configuration.} Parameters held constant across the 96-run factorial experiment unless noted. Factorial factors indicated with $\dagger$.}
\label{tab:supp-config}
\footnotesize
\begin{tabular}{p{5cm}p{7cm}}
\toprule
\textbf{Parameter} & \textbf{Value} \\
\midrule
\multicolumn{2}{l}{\emph{Evolutionary search}} \\
Maximum iterations & 100 \\
Population size (MAP-Elites) & 60 \\
Number of islands & 4 \\
Archive size per island & 25 \\
Elite selection ratio & 0.3 \\
Exploitation ratio & 0.7 \\
Allow full rewrites & True \\
Diff-based evolution & False \\
Number of top programs in prompt & 3 \\
Checkpoint interval & 5 iterations \\
\midrule
\multicolumn{2}{l}{\emph{LLM configuration}} \\
API provider & OpenRouter \\
Primary model$^\dagger$ & GPT-5 or Gemini 3 Flash Preview \\
Primary model weight & 0.80 \\
Secondary model & Claude Sonnet 4 \\
Secondary model weight & 0.20 \\
Temperature & 0.7 \\
Top-$p$ & 0.95 \\
Max output tokens & 8{,}192 \\
API timeout & 600\,s \\
\midrule
\multicolumn{2}{l}{\emph{Evaluation pipeline}} \\
Cascade evaluation & True \\
Stage 1 threshold ($\theta_1$) & 0.001 \\
Parallel workers (stage 2 search) & 1 \\
Parallel workers (final held-out evaluation) & 4 \\
Evaluation timeout per program & 1{,}800\,s \\
\midrule
\multicolumn{2}{l}{\emph{Dataset-specific}} \\
ACIC 2022: training / held-out replicates & 75 / 75 (50:50) or 30 / 120 (20:80), from a 150-replicate working subset sampled from 300 total \\
ACIC 2022: target coverage & 0.90 \\
IHDP: training / held-out replicates & 12 / 13 (50:50) or 5 / 20 (20:80), from a 25-replicate working subset sampled from 50 total \\
ACIC 2016: training / held-out replicates & 50 / 50 (50:50) or 20 / 80 (20:80), from a 100-replicate working subset sampled from 200 total \\
LaLonde: training / held-out replicates & 25 / 25 (50:50) or 10 / 40 (20:80), from a 50-replicate working subset sampled from 100 total \\
\midrule
\multicolumn{2}{l}{\emph{Factorial factors$^\dagger$}} \\
Datasets & ACIC 2022, ACIC 2016, IHDP, LaLonde \\
LLM (primary) & GPT-5, Gemini 3 Flash Preview \\
Regularization $\lambda$ & 0.2 (small), 1.0 (moderate), 5.0 (large) \\
Train--test split & 50:50, 20:80 \\
Independent replicates per cell & 2 \\
\midrule
\multicolumn{2}{l}{\emph{Reproducibility}} \\
Deterministic replicate seeds & 20250215, 20250216 \\
Allowed Python packages & pandas, numpy, scikit-learn, statsmodels \\
\bottomrule
\end{tabular}
\end{table}
\FloatBarrier

\subsubsection*{S1.4. Composite scoring functions}
\manualsectionlabel{sec:supp-scoring-functions}{S1.4}

Each benchmark uses a scalarized score so that MAP-Elites selection can trade off primary accuracy against secondary objectives such as ATE calibration or interval coverage.

\textbf{ACIC 2022 (panel data):}
$S_{\text{ACIC}} = {1}/({1 + \log(1 + \text{RMSE}) + \lambda \cdot \log(1 + |\text{Coverage} - 0.9|/0.9)}).$

\textbf{IHDP, ACIC 2016, LaLonde (cross-sectional):}
$S_{\text{ITE}} = {1}/({1 + \log(1 + \overline{\sqrt{\text{PEHE}}}) + \lambda \cdot \log(1 + \overline{|\text{ATE error}|})}),$
where the overlines denote averages across the training replicates evaluated in each iteration and $\sqrt{\text{PEHE}}$ denotes the unit-level ITE RMSE from equation~\ref{eq:pehe-def}. The three levels of $\lambda$---0.2 (small), 1.0 (moderate), and 5.0 (large)---control the relative weight of secondary metrics.

\subsubsection*{S1.5. Implementation and reproducibility details}
\manualsectionlabel{sec:supp-implementation-details}{S1.5}

The notes below summarize the runtime environment and randomization controls used for the full experiment.

\textbf{Software.} The framework is implemented in Python 3.11 using OpenEvolve~\citep{openevolve} v0.2. LLM API calls are routed through OpenRouter. Candidate programs are executed in isolated subprocesses with enforced timeouts.

\textbf{Reproducibility.} Two deterministic replicate seeds (20250215 and 20250216) control benchmark replicate sampling across the factorial grid. Per-run configuration files and evolved programs at every checkpoint are archived in the code repository.

\textbf{LLM call counts and stylized API cost.} Each evolution iteration issues one candidate-generation LLM call, so a 100-iteration run corresponds to 100 generation calls. Under the 0.80/0.20 primary/secondary ensemble weights, the expected composition is 80 primary-model calls and 20 Claude~Sonnet~4 calls per run. The 96-run true-score factorial therefore corresponds to about 9{,}600 generation calls in total (expected 7{,}680 primary and 1{,}920 Claude calls); including the 48 matched proxy runs raises the total to about 14{,}400 calls. If a run uses on average $b_{\mathrm{in}}$ prompt tokens and $b_{\mathrm{out}}$ completion tokens per generation call, then its expected token budget is $100\,b_{\mathrm{in}}$ input tokens and $100\,b_{\mathrm{out}}$ output tokens, and the corresponding API spend is obtained by multiplying those totals by the provider's then-current model-specific per-token prices. For a representative bookkeeping example of 12k input and 2k output tokens per call, this corresponds to about 1.2M input tokens and 0.2M output tokens per run, about 115M/19M tokens over the 96-run main factorial, and about 173M/29M tokens when the 48 proxy runs are included.

\subsection*{S2. Additional benchmark summaries and complexity analyses}
\manualsectionlabel{sec:supp-benchmark-summaries}{S2}

This section collects the supporting tables and figures referenced from the Results section. We first report the full zero-shot benchmark comparisons, then document the exact program-analysis inputs and prompts, summarize code similarity across final evolved programs, and finally show the expanded code-length and methods-length figures that sit behind the main-text complexity discussion.

\subsubsection*{S2.1. Full zero-shot benchmark tables}
\manualsectionlabel{sec:supp-zero-shot-tables}{S2.1}

Supplementary Tables~\ref{tab:ext-acic-zeroshot} and \ref{tab:ext-ihdp-zeroshot} provide the complete zero-shot comparisons referenced in the ACIC~2022 and IHDP results discussion. They contextualize the seed programs from which the evolutionary search began.

\begin{table}[htbp]
\centering
\caption{\textbf{Full zero-shot results on ACIC~2022 (50 replicates).} Seven LLMs evaluated under identical conditions. Each model was given the same structured prompt containing the data schema, causal assumptions, evaluation criteria, and a six-row data preview. Models were allowed up to five attempts to produce a valid program passing a 180\,s warm-up test on a single replicate.}
\label{tab:ext-acic-zeroshot}
\begin{tabular}{lcccc}
\toprule
Model & Attempts & RMSE & Coverage & Method \\
\midrule
Claude Sonnet 4 & 2 & 89.24 & 0.12 & Weighted AIPW w/ RF \\
GPT-5 & 1 & 33.05 & 0.40 & Patient-weighted TWFE DiD \\
GPT-4o & 1 & 39.21 & 0.36 & DiD + regression adjustment \\
Grok 4 Fast & 3 & 89.24 & 0.22 & Diff-in-means w/ bootstrap \\
Qwen3-Coder & 1 & 68.93 & 0.30 & Weighted TWFE (clustered) \\
DeepSeek v3.1 & 2 & 19.44 & 0.02 & Weighted FE w/ clustered SEs \\
Gemini 2.5 Pro & 5 & --- & --- & \emph{No valid program} \\
\bottomrule
\end{tabular}
\end{table}

\begin{table}[htbp]
\centering
\caption{\textbf{Full zero-shot results on IHDP (50 replicates).} Seven LLMs evaluated under identical conditions. Method sketches are determined by inspecting the generated code. Means $\pm$ SE are computed across 50 replicates with ground-truth potential outcomes.}
\label{tab:ext-ihdp-zeroshot}
\begin{tabular}{lccc}
\toprule
Model & Attempts & $\sqrt{\text{PEHE}}$ & $|\text{ATE error}|$ \\
\midrule
Claude Sonnet 4 & 1 & $2.40 \pm 0.55$ & $0.103 \pm 0.013$ \\
GPT-5 & 1 & $5.68 \pm 0.73$ & $0.292 \pm 0.030$ \\
GPT-4o & 1 & $1.86 \pm 0.38$ & $0.104 \pm 0.014$ \\
Grok 4 Fast & 1 & $1.86 \pm 0.38$ & $0.106 \pm 0.014$ \\
Qwen3-Coder & 2 & $3.11 \pm 0.73$ & $0.489 \pm 0.141$ \\
DeepSeek v3.1 & 2 & $2.02 \pm 0.34$ & $0.103 \pm 0.018$ \\
Gemini 2.5 Pro & 5 & --- & --- \\
\bottomrule
\end{tabular}
\end{table}
\FloatBarrier

\subsubsection*{S2.2. Program-analysis inputs and prompt summaries}
\manualsectionlabel{sec:supp-program-analysis}{S2.2}

This subsection documents the code pool and prompt templates behind the program-analysis results in Fig.~\ref{fig:structure}, Table~\ref{tab:solution_families}, and the complexity figures below. We analyzed one final code snippet per run: 28 zero-shot baselines, 96 true-evolved finals, and 48 proxy-evolved finals from the main 20:80 analysis split. All embedding-based figures and tables in the main text use OpenAI \texttt{text-embedding-3-large}. We reran the same within- versus between-dataset and nearest-reference summaries with \texttt{Qwen3-Embedding-0.6B} as a sensitivity analysis; the alternative embedding space was more compressed, but it preserved the same qualitative ordering and dataset-level conclusions.

\begin{table}[htbp]
\centering
\caption{\textbf{Program-analysis inputs and prompt summaries.} Exact code sources and shared prompt templates for the program-analysis pipeline.}
\label{tab:ext-program-analysis}
\begingroup
\footnotesize
\setlength{\tabcolsep}{4pt}
\renewcommand{\arraystretch}{1.22}
\begin{tabular*}{\linewidth}{@{\extracolsep{\fill}}p{1.55cm}p{2.0cm}p{8.2cm}}
\toprule
Component & Models / inputs & Retrieval rule or shared prompt \\
\midrule
Program pool & 28 baselines, 96 true-evolved finals, 48 proxy-evolved finals & One final snippet per run from the zero-shot baseline pool, the true-score final-program pool, and the proxy-score final-program pool. The exact 172-program manifest is archived alongside the analysis outputs. \\
\rowcolor{TableStripe}
Translation & GPT-5, Gemini~2.5~Pro, Claude~Sonnet~4.5 & Shared system and user template shown below; each prompt supplied the dataset name, source label, and full program text, and asked for a single manuscript-style methods paragraph. \\
Judge panel & GPT-5, Gemini~2.5~Pro, Claude~Sonnet~4.5 & Shared JSON-returning review template shown below; each prompt asked for family assignment, nearest known method, complexity, pipeline steps, and novel components before cross-model aggregation. \\
\rowcolor{TableStripe}
Embedding sensitivity & OpenAI main; Qwen sensitivity & All manuscript figures and tables use \texttt{text-embedding-3-large}. The same nearest-reference and within/between-program summaries were rerun with \texttt{Qwen3-Embedding-0.6B} only as a sensitivity analysis. \\
\bottomrule
\end{tabular*}
\endgroup
\end{table}

For the official reference library, we also archived the exact commit-pinned upstream example links used to build the seven normalized wrappers (see Supplementary Table~\ref{tab:ext-reference-links}).

\begin{table}[htbp]
\centering
\caption{\textbf{Commit-pinned upstream sources for the official reference library.} Each normalized \texttt{estimate(df)} wrapper used in the reference-matching analysis was adapted from the corresponding maintainer example at the listed commit-pinned URL.}
\label{tab:ext-reference-links}
\begingroup
\footnotesize
\setlength{\tabcolsep}{4pt}
\renewcommand{\arraystretch}{1.18}
\begin{tabular*}{\linewidth}{@{\extracolsep{\fill}}p{1.35cm}p{2.6cm}p{1.45cm}p{4.85cm}p{1.6cm}}
\toprule
Family & Wrapper & Package & Commit-pinned upstream example & Commit \\
\midrule
T-learner & T-Learner (EconML TLearner) & EconML & \url{https://github.com/py-why/EconML/blob/b0404c02be5af1f9a2cbc2bd66d9b47e38c6ebb4/econml/metalearners/_metalearners.py#L54-L62} & \texttt{b0404c0} \\
\rowcolor{TableStripe}
X-learner & X-Learner (EconML XLearner) & EconML & \url{https://github.com/py-why/EconML/blob/b0404c02be5af1f9a2cbc2bd66d9b47e38c6ebb4/econml/metalearners/_metalearners.py#L328-L336} & \texttt{b0404c0} \\
DR-learner & DR-Learner (EconML DRLearner) & EconML & \url{https://github.com/py-why/EconML/blob/b0404c02be5af1f9a2cbc2bd66d9b47e38c6ebb4/econml/dr/_drlearner.py#L542-L555} & \texttt{b0404c0} \\
\rowcolor{TableStripe}
IPW & IPW (zEpid IPTW) & zEpid & \url{https://github.com/pzivich/zEpid/blob/16a0f96f8b2c65df8715085801f21757d1478e1e/zepid/causal/ipw/IPTW.py#L85-L88} & \texttt{16a0f96} \\
OLS & OLS (statsmodels OLS) & statsmodels & \url{https://github.com/statsmodels/statsmodels/blob/184e7f8e5725613ba64cbabc9bf0a33354e1d95e/examples/python/ols.py#L31-L37} & \texttt{184e7f8} \\
\rowcolor{TableStripe}
AIPW & AIPW (zEpid AIPTW) & zEpid & \url{https://github.com/pzivich/zEpid/blob/16a0f96f8b2c65df8715085801f21757d1478e1e/zepid/causal/doublyrobust/AIPW.py#L74-L77} & \texttt{16a0f96} \\
Causal forest & CausalForest (EconML CausalForestDML) & EconML & \url{https://github.com/py-why/EconML/blob/b0404c02be5af1f9a2cbc2bd66d9b47e38c6ebb4/econml/dml/causal_forest.py#L551-L558} & \texttt{b0404c0} \\
\bottomrule
\end{tabular*}
\endgroup
\end{table}

\clearpage
\begin{promptbox}{Code-to-Methods Translation Prompt}
\begin{lstlisting}[breaklines=true, basicstyle=\scriptsize\ttfamily, columns=fullflexible, numbers=none, frame=none]
System:
You convert Python estimation code into a methods paragraph for a scientific
manuscript. Describe the statistical procedure, preprocessing, weighting,
model structure, effect-estimation target, and uncertainty quantification when
present. Describe only what the code actually implements. Do not speculate, do
not mention absent steps, and do not add commentary such as "the code does not
explicitly..." or "there is no..." unless the omission is itself an explicit
design choice encoded in the program. Avoid line-by-line narration, code
syntax, bullet points, or implementation trivia that would not belong in a
methods section. Return plain text only.

User:
Dataset: <dataset>
Program source: <source label>

Translate the following Python code into a single publication-style methods
paragraph.

```python
<full program inserted here>
```
\end{lstlisting}
\end{promptbox}

\clearpage
\begin{promptbox}{LLM-as-a-Judge Aggregation Prompt}
\begin{lstlisting}[breaklines=true, basicstyle=\scriptsize\ttfamily, columns=fullflexible, numbers=none, frame=none]
You are a causal inference expert reviewing a Python program that estimates
individual treatment effects (ITE) and average treatment effects (ATE).

Analyze this program and answer the following questions. Be precise and concise.

```python
<full program inserted here>
```

Respond in EXACTLY this JSON format (no markdown, no extra text):
{
  "algorithm_family": "<primary method family: one of T-Learner, S-Learner,
    X-Learner, DR-Learner, AIPW, IPW, OLS, CausalForest, Matching, Ensemble,
    Other>",
  "sub_methods": ["<specific sub-methods used>"],
  "most_similar_known_method": "<single most similar published method>",
  "similarity_to_known": <integer 1-10>,
  "novel_components": ["<algorithmic innovations not in the standard method>"],
  "novelty_explanation": "<1-2 sentence explanation>",
  "base_learners": ["<ML models used>"],
  "pipeline_steps": ["<ordered algorithm steps>"],
  "complexity_rating": <integer 1-10>
}
\end{lstlisting}
\end{promptbox}
\FloatBarrier

\clearpage
\subsubsection*{S2.3. Code similarity among final programs}
\manualsectionlabel{sec:supp-code-similarity}{S2.3}

Supplementary Table~\ref{tab:ext-similarity} complements the main-text method-family summary by quantifying within- and between-dataset lexical similarity across the final evolved programs.

\begin{table}[htbp]
\centering
\caption{\textbf{Code similarity among final programs.} TF-IDF cosine similarity between pairs of final evolved programs. Within-dataset similarity is consistently higher than between-dataset similarity, confirming genuine dataset-specific adaptation.}
\label{tab:ext-similarity}
\begin{tabular}{lcc}
\toprule
Comparison & Mean $\pm$ SD & $n$ pairs \\
\midrule
Within ACIC 2022   & $0.336 \pm 0.110$ & 276 \\
Within ACIC 2016   & $0.353 \pm 0.173$ & 276 \\
Within IHDP        & $0.415 \pm 0.244$ & 276 \\
Within LaLonde     & $0.331 \pm 0.116$ & 276 \\
\midrule
ACIC 2022 vs.\ IHDP      & $0.140 \pm 0.053$ & 576 \\
ACIC 2016 vs.\ IHDP      & $0.352 \pm 0.186$ & 576 \\
ACIC 2022 vs.\ LaLonde   & $0.148 \pm 0.063$ & 576 \\
\bottomrule
\end{tabular}
\end{table}
\FloatBarrier

\subsubsection*{S2.4. Program and methods-length summaries}
\manualsectionlabel{sec:supp-program-lengths}{S2.4}

\begin{figure}[htbp!]
\centering
\includegraphics[width=\textwidth]{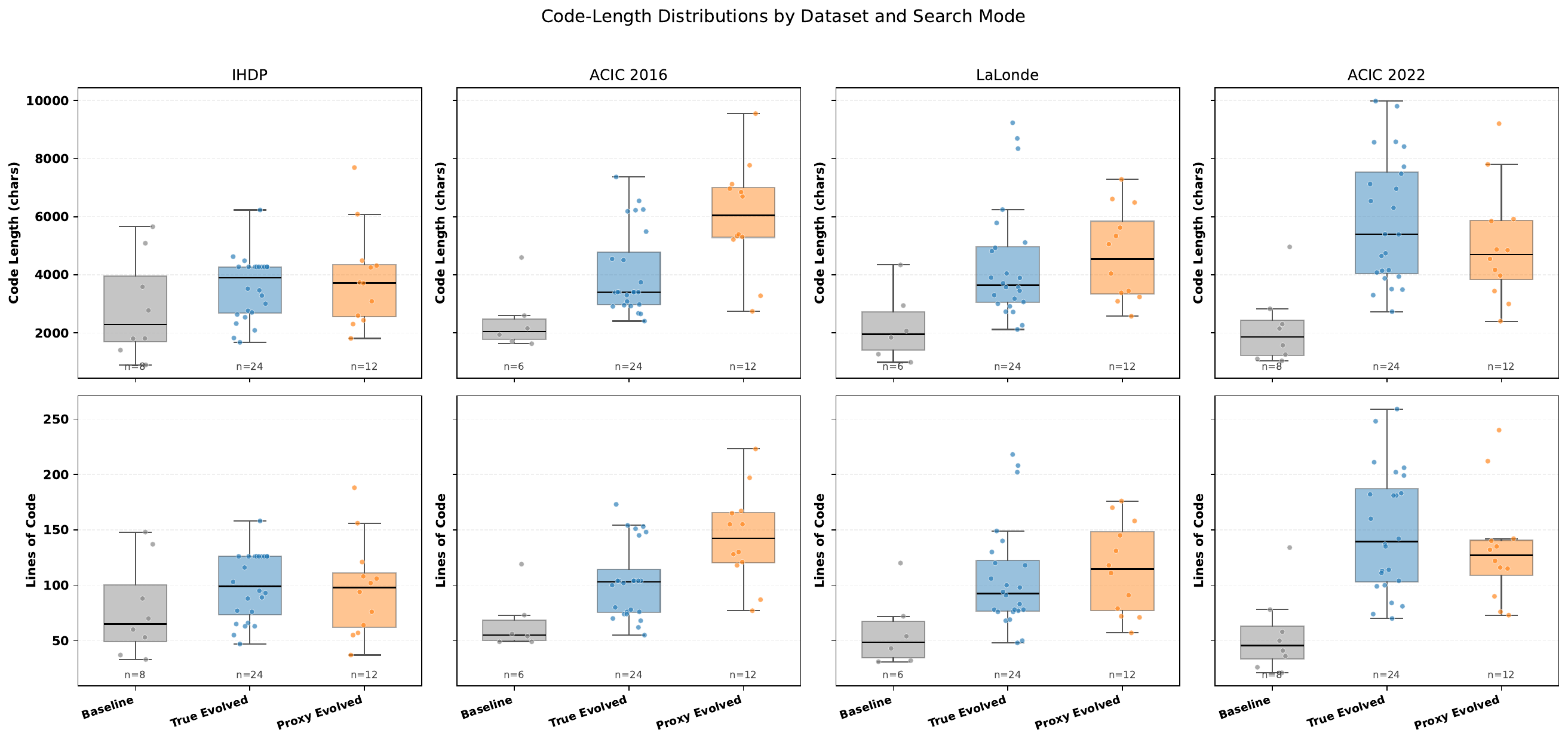}
\caption{\textbf{Code length distributions across datasets and program sources.} Top panels show character count and bottom panels show non-empty lines of code for baseline, true-evolved, and proxy-evolved programs. Across all four benchmarks, evolved programs are consistently longer than the compact baseline programs, with the largest expansions occurring for ACIC~2022 true-evolved code and ACIC~2016 proxy-evolved code.}
\label{fig:ext-code-length}
\end{figure}

\begin{figure}[htbp!]
\centering
\includegraphics[width=\textwidth]{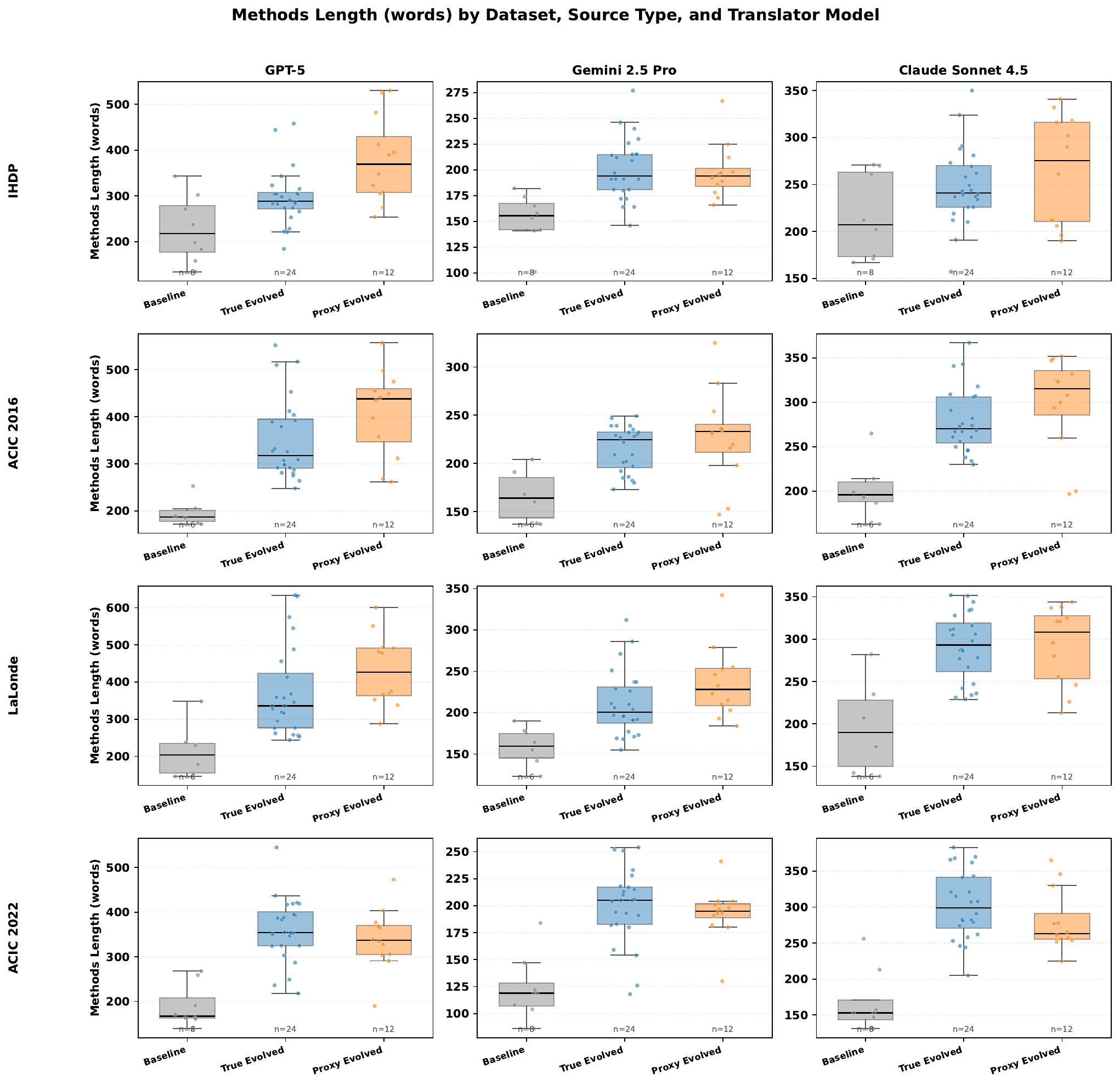}
\caption{\textbf{Lengths of manuscript-style methods paragraphs derived from code.} Each program was translated into a single scientific-manuscript methods paragraph by GPT-5, Gemini~2.5~Pro, and Claude~Sonnet~4.5 using the same unconstrained instruction template. Across datasets and translator models, baseline programs yield the shortest descriptions, while true-evolved and proxy-evolved programs produce substantially longer methods paragraphs.}
\label{fig:ext-method-length}
\end{figure}

\begin{figure}[htbp!]
\centering
\includegraphics[width=\textwidth]{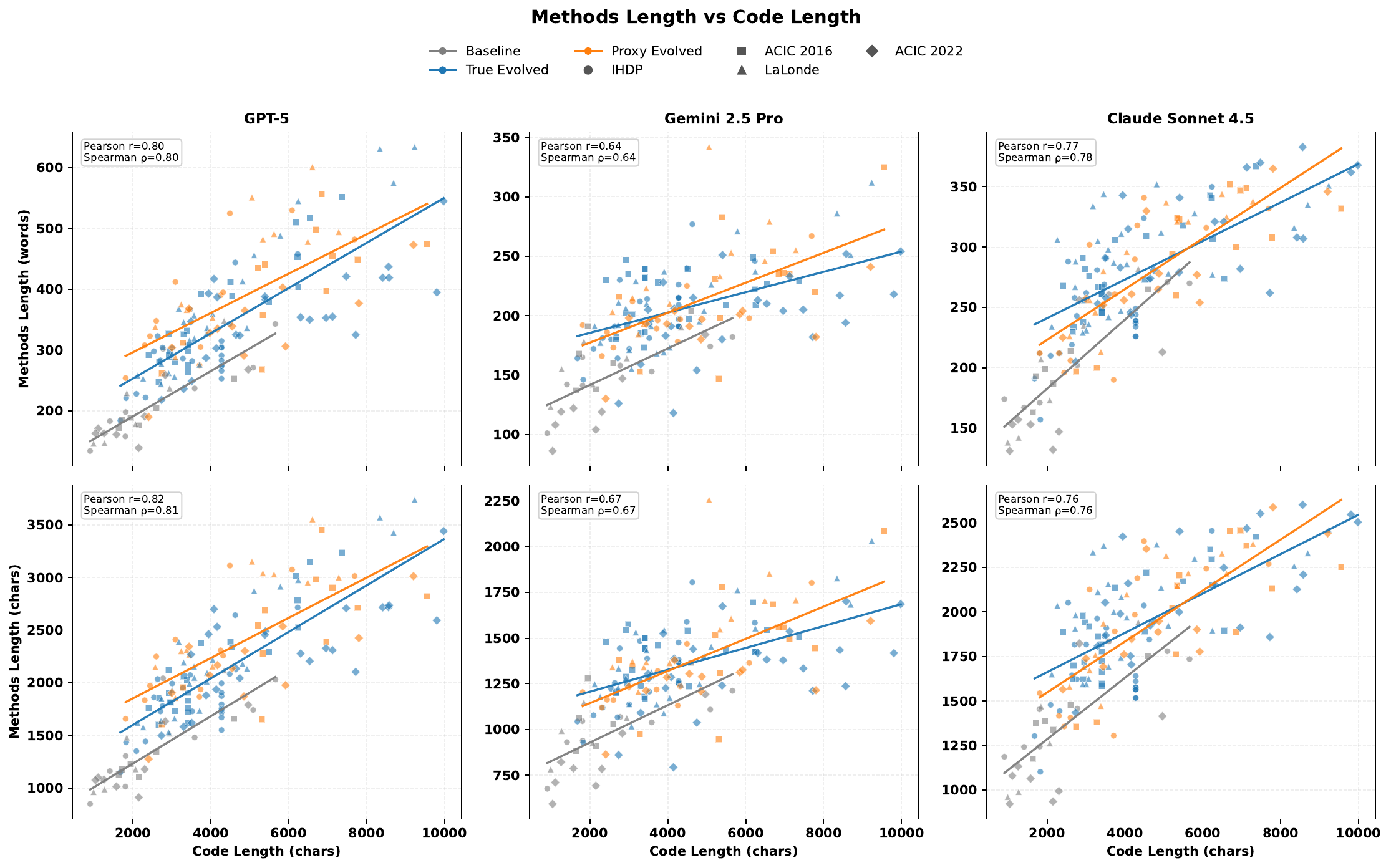}
\caption{\textbf{Translated methods length scales with source-code length.} Each point is one code-to-method translation. Longer source programs tend to yield longer manuscript descriptions across all three translator models, indicating that structural complexity discovered during evolution is reflected in both code size and textual methodological detail.}
\label{fig:ext-method-corr}
\end{figure}

\begin{figure}[htbp!]
\centering
\includegraphics[width=\textwidth]{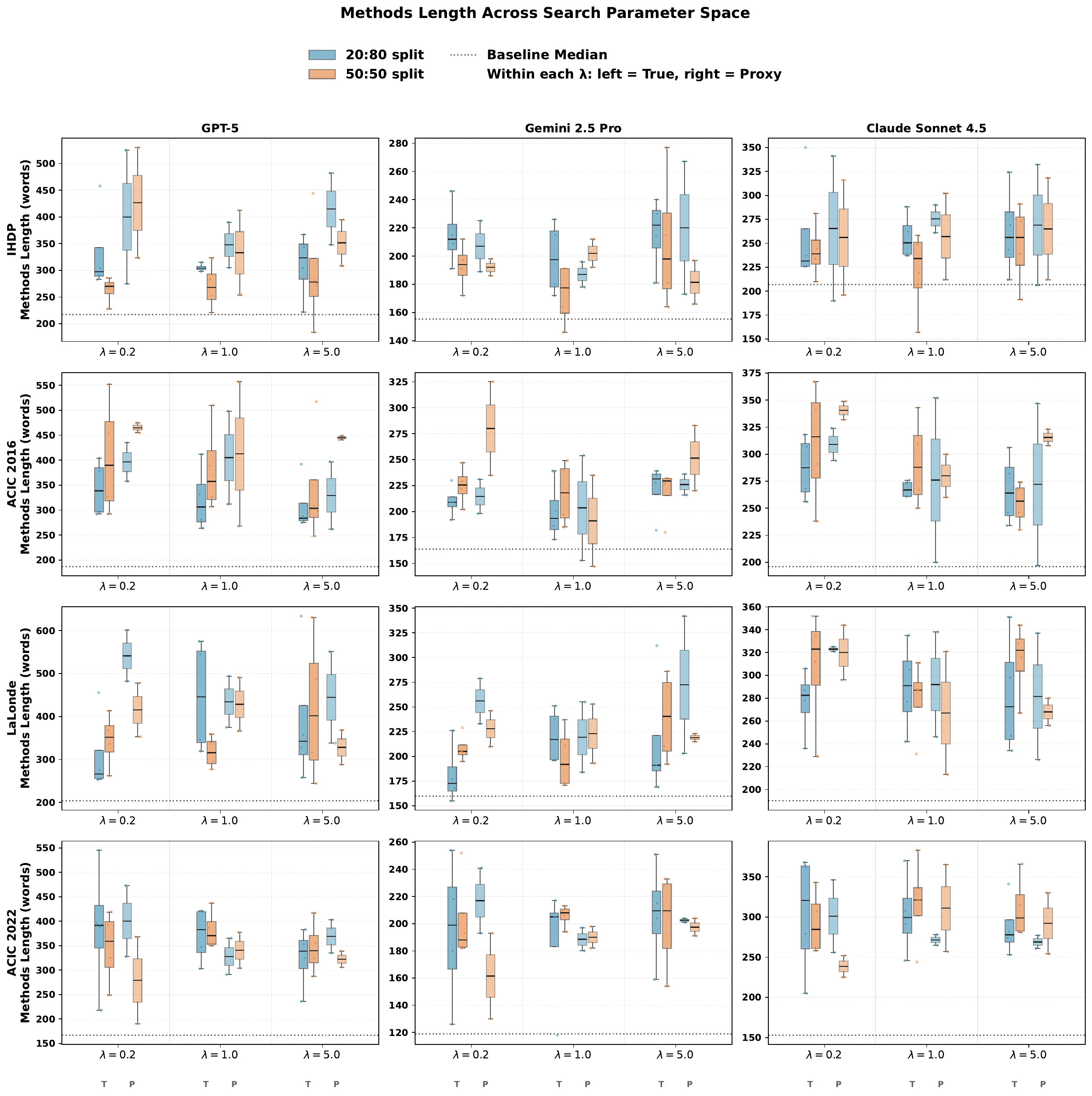}
\caption{\textbf{Methods-paragraph length across the evolutionary parameter space.} Length distributions are stratified by dataset, translator model, source class, train--test split, and regularization level. Within each $\lambda$ group, the left pair shows true-evolved programs and the right pair shows proxy-evolved programs; box color denotes the 20:80 versus 50:50 split. Source effects dominate split and regularization effects, although GPT-5 proxy-derived descriptions are longest under small $\lambda$.}
\label{fig:ext-method-param}
\end{figure}

Supplementary Figs.~\ref{fig:ext-code-length}, \ref{fig:ext-method-length}, \ref{fig:ext-method-corr}, and \ref{fig:ext-method-param} expand the compact complexity panels in Fig.~\ref{fig:baseline}. They show the full distributions of source-code length and manuscript-style methods length across datasets, translator models, and evolutionary settings.

\FloatBarrier

\subsection*{S3. Sensitivity and proxy evaluation details}
\manualsectionlabel{sec:supp-sensitivity-proxy}{S3}

This section collects the supporting analyses behind the design-sensitivity and proxy-evaluation discussion in the Results section. We first document the held-out split sensitivity that is intentionally kept out of the main text, then decompose the regularization-weight comparison into its component metrics, and finally give the technical details and validation summaries for the proxy objectives.

\subsubsection*{S3.1. Held-out split sensitivity}
\manualsectionlabel{sec:supp-heldout-sensitivity}{S3.1}

\begin{figure}[htbp!]
\centering
\includegraphics[width=\textwidth]{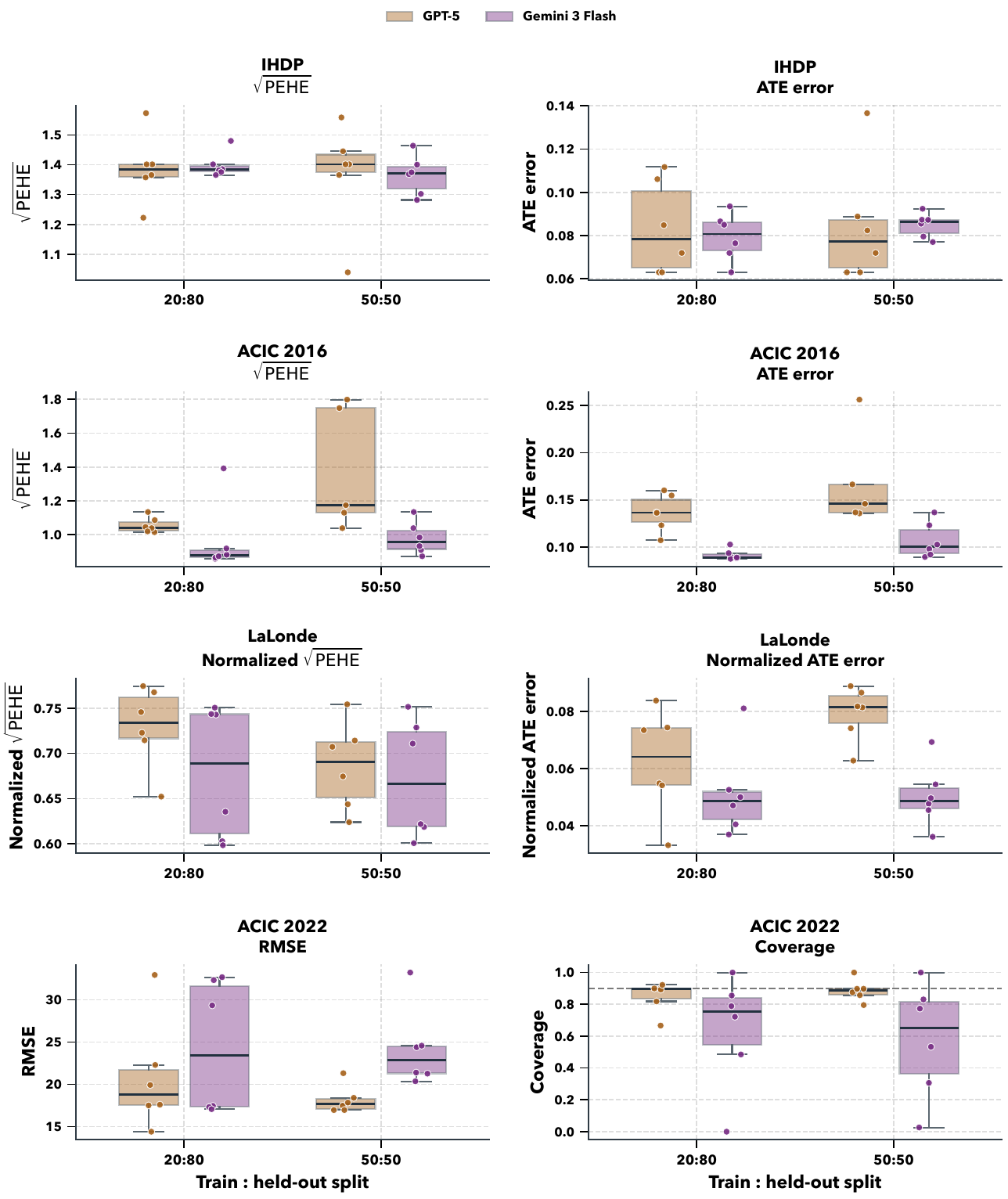}
\caption{\textbf{Held-out sensitivity to the training split.} Each panel compares the prespecified 20:80 main-analysis runs with matched 50:50 reruns, pooling over regularization settings and replicate seeds within each split--model cell; box color denotes the primary model. Columns show the held-out metrics reported in the main text and tables: $\sqrt{\mathrm{PEHE}}$ and ATE error for IHDP and ACIC~2016, normalized $\sqrt{\mathrm{PEHE}}$ and normalized ATE error for LaLonde, and RMSE plus empirical coverage for ACIC~2022. The dashed line in the ACIC~2022 coverage panel marks nominal 90\% coverage. The similar distributions across split settings motivate anchoring the main text on 20:80 alone.}
\label{fig:supp-split-sensitivity}
\end{figure}

All main-text figures and tables use the prespecified 20:80 split. Supplementary Fig.~\ref{fig:supp-split-sensitivity} shows the matched 50:50 reruns for true-score evolution, pooling over regularization settings and replicate seeds within each split--model cell. Across datasets, the held-out metric distributions are similar under 20:80 and 50:50, with split-to-split shifts smaller than the benchmark-level differences emphasized in the main text. We therefore use 20:80 as the single anchor split in the Results section and treat 50:50 as a robustness check.

\FloatBarrier

\subsubsection*{S3.2. Component-wise sensitivity to the regularization weight $\lambda$}
\manualsectionlabel{sec:supp-lambda-sensitivity}{S3.2}

\begin{figure}[htbp!]
\centering
\includegraphics[width=\textwidth]{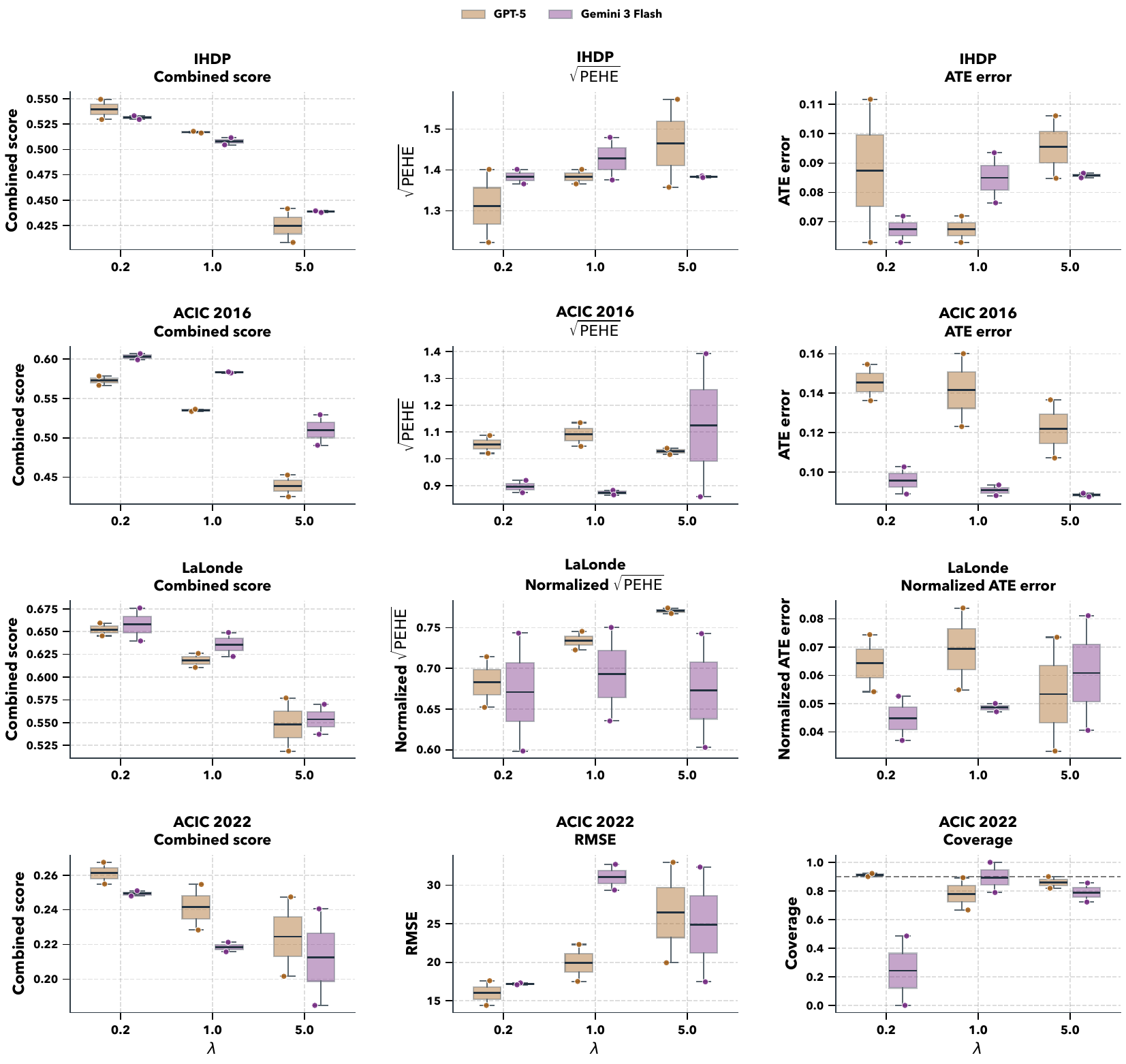}
\caption{\textbf{Sensitivity to $\lambda$ across the combined score and its component metrics.} Each panel pools over replicate seeds within a dataset, $\lambda$, and primary-model cell under the main 20:80 split, with box colors denoting the primary model. Columns show the combined score, the primary accuracy metric, and the secondary metric used in the scalarized objective. Because $\lambda$ appears in the combined-score definition, the first column should be interpreted as a $\lambda$-specific search objective rather than a directly comparable cross-$\lambda$ performance scale; the second and third columns provide the valid held-out metric comparison. For IHDP, ACIC~2016, and LaLonde these component metrics are $\sqrt{\mathrm{PEHE}}$ and ATE error (normalized for LaLonde); for ACIC~2022 they are RMSE and empirical coverage. The dashed line in the ACIC~2022 coverage panel marks nominal 90\% coverage.}
\label{fig:supp-lambda-metrics}
\end{figure}

Supplementary Fig.~\ref{fig:supp-lambda-metrics} expands the bottom row of Fig.~\ref{fig:dynamics-lambda} into the combined score and the two component metrics used in each benchmark-specific objective, again under the main 20:80 split. Because $\lambda$ enters the scalarized score definition, only the component-metric columns support direct cross-$\lambda$ comparison of estimator quality.

\FloatBarrier

\subsubsection*{S3.3. DR pseudo-outcome PEHE proxy}
\manualsectionlabel{sec:supp-dr-proxy}{S3.3}

For heterogeneous treatment effect estimation (IHDP, ACIC~2016, LaLonde-Norm), we construct cross-fitted doubly robust (DR) pseudo-outcomes following \citet{kennedy2024semiparametric}:

\begin{enumerate}
\item Split the data into $K = 2$ folds.
\item For each fold $k$, fit nuisance models on the complement $\{1,\ldots,K\}\setminus k$: outcome models $\hat{\mu}_t(x) = \mathbb{E}[Y \mid X=x, T=t]$ using both Ridge and histogram gradient-boosted regressors; propensity model $\hat{e}(x) = P(T=1 \mid X=x)$ using logistic regression with clipping at $[0.05, 0.95]$; and a main-effect regression $\hat{m}(x)=\mathbb{E}[Y\mid X=x]$ using Ridge for the R-loss term.
\item Construct the DR pseudo-outcome for each unit $i$ in fold $k$:
$\tilde{\tau}_i = \hat{\mu}_1(x_i) - \hat{\mu}_0(x_i) + {T_i(Y_i - \hat{\mu}_1(x_i))}/{\hat{e}(x_i)} - {(1-T_i)(Y_i - \hat{\mu}_0(x_i))}/({1-\hat{e}(x_i)}).$
\item Proxy ITE fit:
\[
\widehat{\mathrm{RMSE}}_{\tau,\text{proxy}}
=
\left(N^{-1}\sum_i (\hat{\tau}_i - \tilde{\tau}_i)^2\right)^{1/2}.
\]
Proxy ATE: $\hat{\tau}_{\text{AIPW}} = N^{-1}\sum_i \tilde{\tau}_i$.
\item We also compute a normalized R-loss term,
\[
R_{\text{loss,norm}}
=
\frac{N^{-1}\sum_i\left(Y_i - \hat{m}(x_i) - (T_i-\hat{e}(x_i))\hat{\tau}_i\right)^2}
{\widehat{\mathrm{Var}}(Y) + 10^{-8}}.
\]
The implemented ITE proxy objective is
\[
\mathcal{L}_{\text{ITE-proxy}} =
0.5\,\phi(\overline{\widehat{\mathrm{RMSE}}_{\tau,\text{proxy}}}) +
0.3\,\lambda_{\text{ATE}}\,\phi(\overline{|\hat{\tau} - \hat{\tau}_{\text{AIPW}}|}) +
0.2\,\phi(\overline{R_{\text{loss,norm}}}),
\]
with $S_{\text{ITE-proxy}} = 1/(1 + \mathcal{L}_{\text{ITE-proxy}})$ and $\phi(x) = \log(1+x)/(1+\log(1+x))$. As in Section~\ref{sec:supp-scoring-functions}, the overline denotes averaging across the training replicates used during search.
\end{enumerate}

\subsubsection*{S3.4. DR-DID coverage proxy (ACIC 2022)}
\manualsectionlabel{sec:supp-drdid-proxy}{S3.4}

For ACIC~2022's panel data setting, we developed a DR difference-in-differences pseudo-target. We fit outcome models for control units in the pre and post periods, fit a propensity model for treatment assignment (clipped to $[0.05, 0.95]$), and construct the DR-DID estimate per replicate. The most reliable proxy signal was whether a candidate estimator's confidence interval covered this DR-DID target. To prevent trivial high-coverage solutions based on overly wide intervals, we also tracked raw interval width and, when enough replicates were available, a secondary width-calibration term
\[
g_{\text{width}} =
\frac{\left|\overline{W} - 2 z_{0.95}\,\widehat{\mathrm{sd}}(\hat{\tau}_{\text{DR-DID}})\right|}
{2 z_{0.95}\,\widehat{\mathrm{sd}}(\hat{\tau}_{\text{DR-DID}})},
\]
which compares the mean reported interval width with the nominal 90\% width implied by the cross-replicate spread of the DR-DID targets. The implemented objective is
\[
\mathcal{L}_{\text{ACIC-proxy}} =
0.6\,\phi(\overline{|\hat{\tau} - \hat{\tau}_{\text{DR-DID}}|}) +
0.25\,\lambda_{\text{cov}}\widetilde{C} +
0.15\,\phi(\overline{W}),
\]
where $\widetilde{C}$ is the hit-rate gap alone when too few replicates are available for calibration, and otherwise a 70:30 blend of hit-rate centering and cross-replicate width calibration. Across 76 checkpoint reevaluations, the DR-DID hit-rate component achieved $\rho = 0.99$ correlation with true coverage. We therefore treat DR-DID hit-rate as the strongest single ACIC proxy signal. Width calibration and raw interval width were nevertheless retained in the implemented scalarized objective as secondary regularizers against degenerate interval behavior, rather than as the main validated proxy component.

\subsubsection*{S3.5. Proxy validation summary}
\manualsectionlabel{sec:supp-proxy-validation}{S3.5}

Supplementary Table~\ref{tab:supp-proxy-validation} summarizes the alignment between the proxy objectives used during search and the true held-out metrics used for the main evaluation.

\begin{table}[htbp]
\centering
\caption{\textbf{Proxy metric validation.} For IHDP, LaLonde-Norm, and ACIC~2016, values report Pearson correlation between matched final-run proxy combined scores and true combined scores. For ACIC~2022, values report Spearman correlation between the DR-DID hit-rate component and true coverage across checkpoint reevaluations.}
\label{tab:supp-proxy-validation}
\footnotesize
\begin{tabular}{lccc}
\toprule
Dataset & Correlation & $n$ & $p$-value \\
\midrule
IHDP & 0.97 & 4 & 0.03 \\
LaLonde-Norm & 0.77 & 4 & 0.23 \\
ACIC 2016$^\ast$ & 0.50 & 3 & 0.67 \\
ACIC 2022 (DR-DID hit-rate vs.\ true coverage) & 0.99 & 76 & $1.3 \times 10^{-67}$ \\
\bottomrule
\end{tabular}
\vspace{4pt}
\noindent{\footnotesize $^\ast$ACIC~2016 includes one additional cross-fitted ensemble proxy rerun (\texttt{\_cf}) beyond the two standard matched proxy runs.}
\end{table}
\FloatBarrier

\subsection*{S4. Dataset-level evolution trajectories}
\manualsectionlabel{sec:supp-trajectories}{S4}

The four subsections below summarize representative best-run trajectories for each benchmark. Each begins by restating the identification problem faced by the agent, then lists the key transitions along the winning run, and finally describes the resulting evolved estimator in prose.

\subsubsection*{S4.1. ACIC 2022 trajectory}
\manualsectionlabel{sec:supp-acic2022-trajectory}{S4.1}

\paragraph*{Seed program and identification problem.} ACIC~2022 presents a panel data problem: the agent must estimate a patient-weighted SATT from practice-year data with pre/post periods. The seed program is a TWFE-DiD estimator with post-year interactions and cluster-robust WLS, generated by GPT-5 in the zero-shot sweep (RMSE~$=33.1$, coverage~$=0.40$).

\paragraph*{Key transitions.} Supplementary Table~\ref{tab:supp-acic-features} details the feature progression across all new-best checkpoints.

\begin{table}[htbp]
\centering
\caption{\textbf{ACIC~2022 feature progression.} Each row is a new-best checkpoint. The agent progressively discovers DiD logic, pre-mean adjustment, and trend correction.}
\label{tab:supp-acic-features}
\footnotesize
\begin{tabular}{clccccccc}
\toprule
Iter. & RMSE & Cov. & Ridge & Cov.\ enc. & Pre-mean & Trend & Post-only & CI adj. \\
\midrule
0 (init) & 33.1 & 0.40 & --- & --- & --- & --- & --- & --- \\
2  & 31.7 & 0.55 & \checkmark & \checkmark & --- & --- & --- & --- \\
14 & 28.7 & 0.55 & --- & \checkmark & \checkmark & --- & --- & --- \\
16 & 27.1 & 0.75 & --- & \checkmark & \checkmark & --- & --- & \checkmark \\
22 & 18.5 & 0.95 & --- & \checkmark & \checkmark & --- & \checkmark & \checkmark \\
52 & 15.3 & 0.90 & --- & \checkmark & \checkmark & \checkmark & \checkmark & \checkmark \\
88 & 15.2 & 0.90 & --- & \checkmark & \checkmark & \checkmark & \checkmark & \checkmark \\
\bottomrule
\end{tabular}
\end{table}

\paragraph*{Final evolved program.} The best ACIC~2022 program (``DR DiD-RA SATT, cluster-robust'') computes per-practice pre-period weighted means and slopes, differences $Y_{\text{post}} - \bar{Y}_{\text{pre}}$, selects the top 8 covariates by correlation with control-group $\Delta Y$, runs per-year ridge-regularized WLS regression adjustment on controls, computes patient-weighted ATT on residuals, and estimates cluster-robust variance using influence-function contributions with conservative (max of residual vs.\ raw) variance selection and 5\% SE inflation.
\FloatBarrier

\subsubsection*{S4.2. ACIC 2016 trajectory}
\manualsectionlabel{sec:supp-acic2016-trajectory}{S4.2}

\paragraph*{Seed program and identification problem.} ACIC~2016 is a cross-sectional benchmark with 58 covariates and mixed continuous/categorical features. The seed program is an X-learner with cross-fitted gradient-boosted outcome models and constant propensity weighting.

\paragraph*{Key transitions.} Supplementary Table~\ref{tab:supp-acic2016-checkpoints} follows a representative best-run trajectory, in which the agent refines the seed X-learner by replacing constant propensity weighting with a random-forest propensity model and local propensity-weighted blending.

\begin{table}[htbp]
\centering
\caption{\textbf{ACIC~2016 evolution checkpoints.} Best-run trajectory (Gemini, $\lambda=0.2$, split 20:80). The agent refines a simple X-learner into a random-forest propensity-weighted X-learner with local blending.}
\label{tab:supp-acic2016-checkpoints}
\footnotesize
\begin{tabular}{clccc}
\toprule
Iter. & Method & $\sqrt{\text{PEHE}}$ & $|\text{ATE error}|$ & Combined \\
\midrule
Init & X-Learner (GBM, cross-fitted) & 1.225 & --- & baseline \\
Early & X-Learner + RF propensity (local weighting) & 1.028 & 0.105 & 0.579 \\
Final & X-Learner (GBM + RF PS) & 1.028 & 0.105 & 0.579 \\
\bottomrule
\end{tabular}
\end{table}

\paragraph*{Final evolved program.} The best ACIC~2016 program (``X-Learner, GBM + RF PS'') replaces the seed's constant propensity weighting with a random forest propensity model, enabling local propensity-weighted blending $\hat{\tau}(x) = \hat{e}(x)\hat{\tau}_0(x) + (1-\hat{e}(x))\hat{\tau}_1(x)$. It uses \texttt{pd.get\_dummies} to handle the 58 mixed-type covariates, 5-fold cross-fitting for outcome surfaces, and separate second-stage GBM models for treated and control pseudo-outcomes. The learning rate is tuned to 0.07 (vs.\ 0.05 in the seed). A separate high-performing GPT-5 variant, not the trajectory shown in Supplementary Table~\ref{tab:supp-acic2016-checkpoints}, evolved further into a ``Meta-Ensemble (T+DR+DRF, adaptive)'' that combines three base learners---T-learner, DR pseudo-outcome, and DR-smoothed random forest---using inverse-variance meta-weights from cross-validated performance.
\FloatBarrier

\subsubsection*{S4.3. IHDP trajectory}
\manualsectionlabel{sec:supp-ihdp-trajectory}{S4.3}

\paragraph*{Seed program and identification problem.} IHDP is a semi-synthetic dataset with 25 covariates and known response surfaces. The seed program is identical to ACIC~2016: an X-learner with cross-fitted GBM outcome models and constant propensity weighting.

\paragraph*{Key transitions.} Supplementary Table~\ref{tab:supp-ihdp-checkpoints} tracks a representative best-run trajectory, in which the agent augments the seed X-learner with an ensemble backbone, neural treatment-effect models, and adaptive propensity weighting.

\begin{table}[htbp]
\centering
\caption{\textbf{IHDP evolution checkpoints.} Best-run trajectory (GPT-5, $\lambda=0.2$, split 20:80). The agent augments the X-learner with an ensemble backbone, neural network treatment effect models, and adaptive propensity weighting.}
\label{tab:supp-ihdp-checkpoints}
\footnotesize
\begin{tabular}{clcc}
\toprule
Iter. & Method & $\sqrt{\text{PEHE}}$ & $|\text{ATE error}|$ \\
\midrule
Init & X-Learner (GBM, cross-fitted, constant $g$) & 1.613 & 0.116 \\
Early & GBM+Ridge ensemble ($0.7/0.3$) + polynomial features & $\approx$1.4 & $\approx$0.10 \\
Mid & + MLP treatment effect models (TARNet-style) & $\approx$1.3 & $\approx$0.10 \\
Final & TARNet-style X-learner + ensemble + adaptive-$g$ & 1.253 & 0.115 \\
\bottomrule
\end{tabular}
\end{table}

\paragraph*{Final evolved program.} The best IHDP program (``TARNet-style X-learner + ensemble + adaptive-$g$'') makes three key changes from the seed: (1)~\emph{Ensemble outcome models}: cross-fitted 70/30 blends of GBM and Ridge regression replace pure GBM, improving stability on the small IHDP samples. (2)~\emph{Neural network treatment effect models}: the second-stage $\hat{\tau}_1, \hat{\tau}_0$ models use MLPRegressor (hidden layers 50--25, L2 regularization $\alpha=0.01$) with GBM fallback, capturing nonlinear heterogeneity. (3)~\emph{Feature-dependent propensity weighting}: rather than the constant $g = \bar{T}$ used in the seed, the program first computes a per-unit feature score $f_i = p^{-1}\sum_{j=1}^{p}|X^{\mathrm{scaled}}_{ij}|$, the mean absolute value of the standardized covariates for unit $i$, then min-max normalizes $f_i$ across units and sets $g_i = \operatorname{clip}(g + 0.1(f_i - 0.5), 0.05, 0.95)$ for propensity-weighted blending of the X-learner arms. Additionally, the program adds interaction features between the top 5 scaled covariates and a propensity proxy, expanding the feature space from 25 to 30 dimensions.
\FloatBarrier

\subsubsection*{S4.4. LaLonde trajectory}
\manualsectionlabel{sec:supp-lalonde-trajectory}{S4.4}

\paragraph*{Seed program and identification problem.} LaLonde is an observational job-training study with only 8 covariates but severe selection bias between experimental participants and survey-based controls. The seed program is the same X-learner used for IHDP and ACIC~2016.

\paragraph*{Key transitions.} Supplementary Table~\ref{tab:supp-lalonde-checkpoints} follows a representative best-run trajectory in which the agent blends X-learner and R-learner components with AIPW calibration.

\begin{table}[htbp]
\centering
\caption{\textbf{LaLonde evolution checkpoints.} Best-run trajectory (GPT-5, $\lambda=0.2$, split 20:80). The agent blends X-learner and R-learner components with AIPW calibration.}
\label{tab:supp-lalonde-checkpoints}
\footnotesize
\begin{tabular}{clcc}
\toprule
Iter. & Method & $\sqrt{\text{PEHE}}$ & $|\text{ATE error}|$ \\
\midrule
Init & X-Learner (GBM, cross-fitted, constant $g$) & --- & --- \\
Early & + Cross-fitted propensity model + DR pseudo-outcomes & $\approx$1.5 & $\approx$0.20 \\
Mid & X-learner + R-learner blend with AIPW & $\approx$1.4 & $\approx$0.17 \\
Final & CF X+R blend, AIPW-calib (GBR+Ridge) & 1.359 & 0.161 \\
\bottomrule
\end{tabular}
\end{table}

\paragraph*{Final evolved program.} The best LaLonde program (``CF X+R blend, AIPW-calib, GBR+Ridge'') is the most architecturally complex of the four datasets, reflecting the difficulty of observational bias correction. It combines:
\begin{itemize}
\item \emph{Cross-fitted nuisance estimation}: 5-fold cross-fitting for outcome models (GBM, 160 trees, depth 3, subsample 0.8) and propensity scores (logistic regression with $C=1.0$, clipped to $[0.02, 0.98]$).
\item \emph{X-learner component}: standard two-stage X-learner with GBM second-stage models and propensity-weighted blending $\hat{\tau}_X = \hat{e}\hat{\tau}_0 + (1-\hat{e})\hat{\tau}_1$.
\item \emph{R-learner component}: residual-on-residual regression $Z_R = (Y - \hat{\mu}_{\text{mix}})/(T - \hat{e})$ with Ridge regression ($\alpha=1.0$), weighted by $(T-\hat{e})^2$.
\item \emph{Ensemble blending}: 60\% X-learner $+$ 40\% R-learner, calibrated so that $\bar{\hat{\tau}} = \hat{\tau}_{\text{AIPW}}$ via an additive shift.
\item \emph{Winsorization}: ITE predictions clipped at the 1st and 99th percentiles.
\end{itemize}

The evolution of an R-learner component is notable because the R-learner was not present in the seed program and represents a genuinely different identification strategy (residual-based) from the X-learner (imputation-based). The 60/40 blend weight and AIPW calibration were discovered through iterative mutation rather than being pre-specified.
\FloatBarrier

\subsection*{S5. Representative code edits during evolution}
\manualsectionlabel{sec:supp-code-edits}{S5}

This final section reproduces short diffs from major inflection points in the evolutionary runs. The goal is to make the trajectory descriptions above concrete by showing the kinds of code mutations associated with meaningful performance jumps.

Listings~\ref{lst:acic-baseline-iter2}, \ref{lst:acic-iter16-iter22}, \ref{lst:ihdp-seed-best}, \ref{lst:acic2016-seed-best}, and \ref{lst:lalonde-seed-best} show representative mutations for ACIC~2022, IHDP, ACIC~2016, and LaLonde, respectively.

\subsubsection*{S5.1. ACIC 2022: Baseline $\to$ Iteration 2 (ridge regularization)}
\manualsectionlabel{sec:supp-acic2022-edit-ridge}{S5.1}

\begin{lstlisting}[language=diff,caption={ACIC 2022: Ridge regularization and covariate encoding added.},label={lst:acic-baseline-iter2}]
@@
-import statsmodels.api as sm
+import statsmodels.api as sm
+from sklearn.preprocessing import StandardScaler
+from sklearn.linear_model import Ridge

@@
-    req = ['id.practice', 'year', 'Y', 'Z', 'post', 'n.patients']
-    for c in req:
-        if c not in df.columns:
-            raise ValueError(f"Missing required column: {c}")
+    d = df.copy()
+    covariate_cols = [col for col in d.columns if col.startswith(('X','V'))]
+    for col in covariate_cols:
+        if d[col].dtype == 'object' or d[col].dtype.name == 'category':
+            dummies = pd.get_dummies(d[col], prefix=col, drop_first=True,
+                                     dtype=float)
+            d = pd.concat([d, dummies], axis=1)
\end{lstlisting}

\subsubsection*{S5.2. ACIC 2022: Iteration 16 $\to$ 22 (post-period differencing)}
\manualsectionlabel{sec:supp-acic2022-edit-postdiff}{S5.2}

\begin{lstlisting}[language=diff,caption={ACIC 2022: Pivotal mutation introduces DiD differencing.},label={lst:acic-iter16-iter22}]
@@
-    post_years = sorted(d.loc[d['post']==1, 'year'].unique().tolist())
+    pre = d.loc[d['post'] == 0].copy()
+    pre_means = (pre.groupby('id.practice')
+                    .apply(lambda g: np.average(g['Y'],
+                                                weights=g['n.patients']))
+                    .rename('Y_pre_mean').reset_index())
+    d = d.merge(pre_means, on='id.practice', how='left')
+    post = d.loc[(d['post']==1) & np.isfinite(d['Y_pre_mean'])].copy()
+    post['D'] = post['Y'].astype(float) - post['Y_pre_mean'].astype(float)
\end{lstlisting}

\clearpage
\subsubsection*{S5.3. IHDP: Seed $\to$ best (ensemble outcome models + neural treatment effects)}
\manualsectionlabel{sec:supp-ihdp-edit}{S5.3}

\begin{lstlisting}[language=diff,caption={IHDP: Seed X-learner evolves to TARNet-style ensemble.},label={lst:ihdp-seed-best}]
@@
-from sklearn.ensemble import GradientBoostingRegressor
+from sklearn.ensemble import GradientBoostingRegressor
+from sklearn.linear_model import Ridge
+from sklearn.neural_network import MLPRegressor

@@  # Cross-fitted outcome models
-        m0 = GradientBoostingRegressor(**params)
-        m0.fit(X_train[control_idx], y_train[control_idx])
-        mu0_preds[test_idx] = m0.predict(X_test)
+        gbr0 = GradientBoostingRegressor(**gbr_params)
+        ridge0 = Ridge(alpha=1.0, random_state=42)
+        gbr0.fit(Xtr[c_mask], ytr[c_mask])
+        ridge0.fit(Xtr[c_mask], ytr[c_mask])
+        mu0_oof[te] = 0.7 * gbr0.predict(Xte) + 0.3 * ridge0.predict(Xte)

@@  # Second-stage treatment effect models
-    tau0 = GradientBoostingRegressor(n_estimators=100, ...)
-    tau0.fit(X_scaled[control_idx], d0[control_idx])
+    tau1_nn = MLPRegressor(hidden_layer_sizes=(50, 25), alpha=0.01, ...)
+    tau1_nn.fit(X_enhanced[t == 1], d1[t == 1])

@@  # Propensity weighting
-    g = np.mean(treatment)
-    ite_hat = g * tau1_pred + (1 - g) * tau0_pred
+    feature_weight = np.mean(np.abs(X_scaled), axis=1)
+    feature_weight = (feature_weight - feature_weight.min()) / (...)
+    adaptive_g = g + 0.1 * (feature_weight - 0.5)
+    adaptive_g = np.clip(adaptive_g, 0.05, 0.95)
+    ite_hat = adaptive_g * tau1_pred + (1 - adaptive_g) * tau0_pred
\end{lstlisting}

\clearpage
\subsubsection*{S5.4. ACIC 2016: Seed $\to$ best (local propensity weighting)}
\manualsectionlabel{sec:supp-acic2016-edit}{S5.4}

\begin{lstlisting}[language=diff,caption={ACIC 2016: Constant propensity replaced by RF propensity model.},label={lst:acic2016-seed-best}]
@@
-from sklearn.ensemble import GradientBoostingRegressor
+from sklearn.ensemble import GradientBoostingRegressor, RandomForestClassifier

@@  # Propensity score: global mean -> local RF estimate
-    g = np.mean(treatment)
-    g_pred = np.full(len(df), g)
-    ite_hat = g_pred * tau1_pred + (1 - g_pred) * tau0_pred
+    ps_model = RandomForestClassifier(n_estimators=100, max_depth=5,
+                                       min_samples_leaf=10, random_state=42)
+    ps_model.fit(X, t)
+    ps = np.clip(ps_model.predict_proba(X)[:, 1], 0.05, 0.95)
+    ite_hat = ps * tau0_x + (1 - ps) * tau1_x
\end{lstlisting}

\subsubsection*{S5.5. LaLonde: Seed $\to$ best (R-learner + AIPW calibration)}
\manualsectionlabel{sec:supp-lalonde-edit}{S5.5}

\begin{lstlisting}[language=diff,caption={LaLonde: X-learner augmented with R-learner and AIPW calibration.},label={lst:lalonde-seed-best}]
@@  # New: R-learner component (not in seed)
+    mu_mix = ehat * mu1 + (1 - ehat) * mu0
+    y_res = y - mu_mix
+    t_res = t - ehat
+    z_r = np.where(np.abs(t_res) > 1e-3, y_res / t_res, 0.0)
+    w_r = np.clip(t_res ** 2, 1e-4, None)
+    r_model = Ridge(alpha=1.0, random_state=42)
+    r_model.fit(Xs, z_r, sample_weight=w_r)
+    tau_r = r_model.predict(Xs)

@@  # New: Ensemble blend + AIPW calibration (not in seed)
+    ite_hat = 0.6 * tau_x + 0.4 * tau_r
+    shift = ate_dr - float(np.mean(ite_hat))
+    ite_hat = ite_hat + shift  # calibrate to AIPW ATE
\end{lstlisting}

\end{document}